\documentclass{article}
\usepackage{times}
\usepackage[utf8]{inputenc} 
\usepackage[T1]{fontenc}    
\usepackage{url}            
\usepackage{booktabs}       
\usepackage{nicefrac}       
\usepackage{microtype}      
\usepackage{adjustbox}      
\usepackage{subcaption}
\usepackage{wrapfig}
\usepackage[font=small,labelfont=bf]{caption}
\usepackage[
  separate-uncertainty = true,
  multi-part-units = repeat
]{siunitx}

\makeatletter
\let\NAT@parse\undefined
\makeatother

\usepackage{algorithm}
\usepackage[noend]{algpseudocode}
\algrenewcommand\algorithmicindent{1em}
\algrenewcommand{\algorithmiccomment}[1]{%
\bgroup\hskip2em\textcolor{ourdarkgreen}{//~\textsl{#1}}\egroup}

\usepackage{enumitem}

\usepackage{amsfonts}       
\usepackage{amsmath}       
\usepackage{amssymb}
\usepackage{xspace}
\usepackage{multirow}
\usepackage{dblfloatfix}
\usepackage{placeins}
\usepackage[]{footmisc}

\usepackage{xr-hyper}
\usepackage[linkcolor=ourblue, citecolor=ourgreen, colorlinks=true, urlcolor=ourblue]{hyperref}
\usepackage[capitalise]{cleveref} 

\makeatletter
\DeclareRobustCommand\onedot{\futurelet\@let@token\@onedot}
\def\@onedot{\ifx\@let@token.\else.\null\fi\xspace}
\makeatother

\newcommand{\RNum}[1]{\uppercase\expandafter{\romannumeral #1\relax}}

\makeatletter
\newcommand*{\addFileDependency}[1]{
  \typeout{(#1)}
  \@addtofilelist{#1}
  \IfFileExists{#1}{}{\typeout{No file #1.}}
}
\makeatother

\usepackage{xcolor}
\definecolor{ourblue}{rgb}{0.368,0.507,0.71}
\definecolor{ourorange}{rgb}{0.881,0.611,0.142}
\definecolor{ourgreen}{rgb}{0.56,0.692,0.195}
\definecolor{ourred}{rgb}{0.923,0.386,0.209}
\definecolor{ourviolet}{rgb}{0.528,0.471,0.701}
\definecolor{ourbrown}{rgb}{0.772,0.432,0.102}
\definecolor{ourlightblue}{rgb}{0.364,0.619,0.782}
\definecolor{ourdarkgreen}{rgb}{0.572,0.586,0.}

\definecolor{ourcyan2}{rgb}{0.125,0.722,0.804}
\definecolor{ourred2}{rgb}{0.863,0.184,0.047}
\definecolor{ouryellow2}{cmyk}{0,0.16,1.0,0.07}
\definecolor{ourviolet2}{cmyk}{0.55,0.56,0,0.47}
\definecolor{ourorange2}{cmyk}{0,0.46,0.89,0.11}

\usepackage[preprint]{corl_2026} 
\usepackage{wrapfig}
\usepackage{caption}

\title{\LARGE \bf
Belt-Finger: An Affordable Soft Belt-Driven Gripper for Dexterous In-Hand Manipulation
}
\newcommand{\method}{Belt-Finger\xspace}

\author{Boya Zhang$^{1}$, Andreas Zell$^{1}$ and Georg Martius$^{1,2}$
\thanks{$^{1}$Department of Computer Science, University of Tübingen, Germany
        {\tt\small firstname.surname@uni-tuebingen.de}}%
\thanks{$^{2}$Max Planck Institute for Intelligent Systems, Germany
        {\tt\small surname@is.mpg.de}}%
}

\newcommand{\aroll}{\ensuremath{a_{\mathrm{roll}}}}
\newcommand{\aopening}{\ensuremath{a_{\mathrm{opening}}}}
\newcommand{\C}{\ensuremath{\mathcal C}}

\begin{document}
\maketitle


\begin{abstract}
Parallel-jaw grippers are the default manipulator choice in robotics because they are simple, robust, and inexpensive. Their limited in-hand mobility, however, often forces large arm motions and restricts dexterous manipulation in confined workspaces.
We present a parallel-gripper upgrade: a double-soft-belt-based finger module that preserves standard opening/closing while adding three in-hand degrees of freedom (DoF): translation, pitch, and roll.
The mechanism is deliberately kept simple and engineered for inexpensive manufacturing and straightforward integration, preserving the reliability and precise control of traditional parallel grippers while greatly broadening the range of manipulation capabilities.
To demonstrate the utility of the added DoFs, we integrate the gripper in two control pipelines. First, we adapt a model predictive controller for in-hand manipulation of known objects. Second, we introduce a lightweight teleoperation interface that enables simultaneous control of the robot arm and gripper (10 DoFs total) with minimal hardware. Across a suite of challenging manipulation tasks executed via teleoperation, MPC, and trained policies, the proposed gripper consistently improves dexterity and task feasibility compared to a conventional parallel gripper.\looseness-1

\end{abstract}


\begin{center}
    \centering
   \includegraphics[width=.8\linewidth]{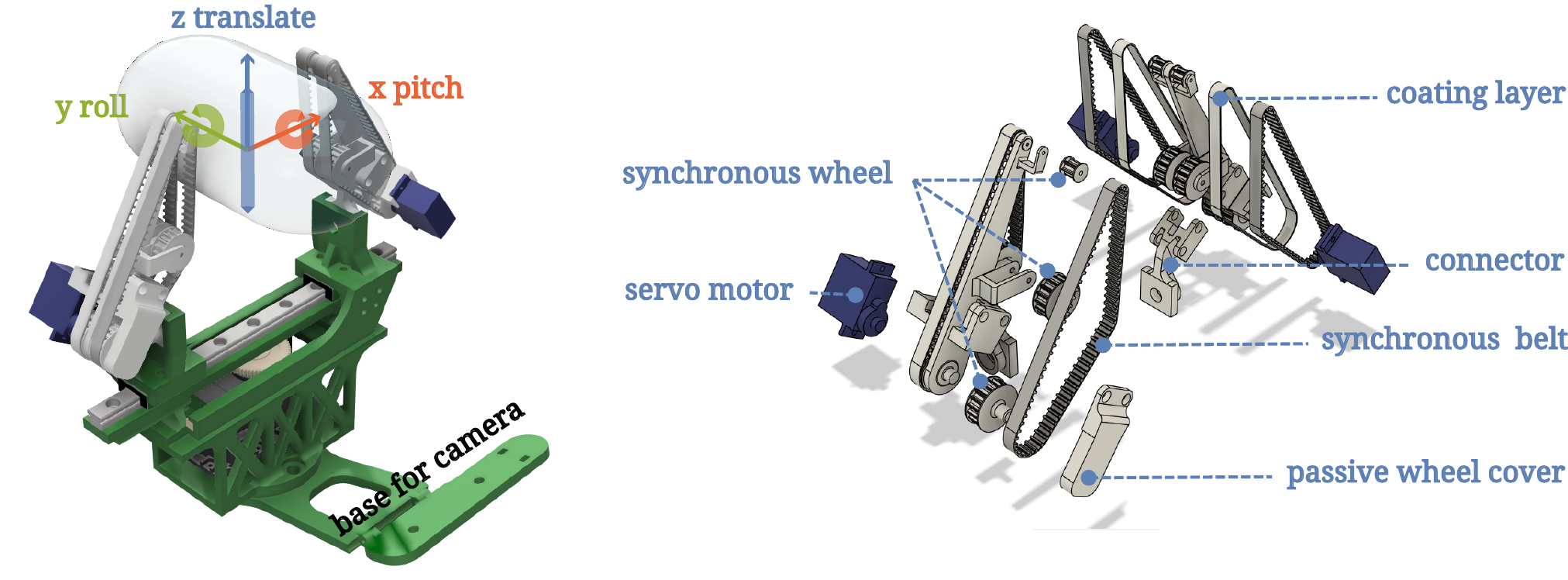}
  \captionof{figure}{{\bf \method design.} The left side shows the gripper made of two belt-enhanced jaws grasping a dummy object. Three colors show the in-hand manipulation DoFs in the local frame. The right side shows the exploded view of the \method. There are three active wheels, each driven by a servo motor, with a coating layer outside of each belt to increase friction forces.}
  \label{fig:gripper-render}
\end{center}%

\section{INTRODUCTION}

The applicability and success of robotic manipulation are largely dependent on the qualities of the end-effector.
Among various options, the parallel gripper, due to its simple structure, remains the most common choice and is widely used in both industrial and household applications.
Current alternatives that provide in-hand manipulation are robotic hands, which come with a significant increase in manufacturing costs and control complexity.
To deal with this dilemma, we present a drop-in replacement for the parallel gripper to enhance its manipulation capabilities by incorporating three actively driven belts mounted on the fingers.
To highlight its versatility, we demonstrate effective control of the gripper across diverse paradigms. Specifically, we utilize an intuitive teleoperation system for human demonstrations (e.g., card insertion), an iCEM-based MPC for precise, zero-shot manipulation, and finetuned Vision-Language-Action (VLA) policies that bridge high-level reasoning with low-level dexterity. Ultimately, these evaluations show our design outperforms classic parallel grippers in usability, efficiency, and user satisfaction.

We make the following contributions:
\begin{itemize}[nosep]
    \item \method{}: A 3D-printable, VLA-friendly, soft-belt-driven gripper for superior complex object manipulation.
    \item A teleoperation system designed for intuitive human control to perform demonstrations.
    \item A simplified model for performing stochastic MPC for in-hand manipulation.
\end{itemize}

\section{Related works}\label{sec:related-work}
\paragraph{Grippers and hands  with in-hand manipulation ability}
In-hand manipulation abilities are clearly desired to solve increasingly complex tasks. Many robotic hands have been proposed, such as
the Shadow hand \cite{ShadowRobot} involving 20 actively controllable joints.
Similar works, e.g., the LEAP Hand \cite{shaw2023leap} and the Everyday finger \cite{ornelas2024everyday}, differ in implementation details but also involve long kinematics chains for fingers and rigid links. 
Hands have many advantages, but their downsides are high costs, robustness challenges, and the required complex control strategies.
On the other side, soft grippers, e.g. \cite{mack2023soft,teeple2021role} with multiple continuous bendable fingers are developed.
While intriguing, accurate control and robustness remain challenging. 

\begin{figure*}[ht] 
  \centering
  
  \begin{minipage}[t]{0.58\linewidth}
    \centering
    \vspace{0pt} 
    
    \captionof{table}{\textsc{Comparison of end effectors}} 
    \label{tab:comprison-of-grippers}
    \vspace{.2em}
    
    \footnotesize
    \adjustbox{max width=\linewidth}{
      \begin{tabular}{@{}l|c|c|c|c|c@{}}
      \toprule
      \multirow{ 2}{*}{\textbf{Type}} & \textbf{In-hand} & \textbf{Actuators} & \textbf{Kinematic} & \textbf{Contact} & \textbf{Portable} \\
      & \textbf{DoFs} & \textbf{Number} & \textbf{Chain} & \textbf{Type} & \textbf{Finger}\\
      \midrule
      Shadow hand \cite{ShadowRobot} & 3T+3R & 20 & >3  & rigid  & ---\\\midrule
      Roller Grasper \cite{yuan2020design} & 3T+3R & 9 & 3 & rigid  & no \\
      Reconfigurable Gripper \cite{11082440} & 1T+2R & 9 & 2 & hybrid & no \\
      DexGrip \cite{wang2025dexgrip} & 1T+3R  & 8 & 2  & hybrid  & no \\
      GRIP-tape \cite{he2023grip} & 2T+1R  & 7 & 2 & rigid  & no \\
      VF finger \cite{spiers2018variable} & 2T+1R & 4 & 2 & rigid & no \\
      TRRG \cite{yuan2025tactile} & 1T+2R & 6 & 3  & hybrid & yes\\
      DTactive \cite{xu2024dtactive} & 1T+2R & 5 & 3  & hybrid  & yes \\
      Band-Based Gripper \cite{xiang2024adaptive} & 2T+1R  & 4 & 2  & soft  & no\\
      Everyday Finger \cite{ornelas2024everyday} & 1T+1R & 4 & 3 & hybrid & yes\\
      The Velvet Fingers \cite{tincani2013implementation} & 1T+1R  & 3 & 2/3  & rigid & no \\
      RoTipBot \cite{jiang2025rotipbot} & 1T+1R  & 3 & 2  & hybrid & yes\\
      BOP \cite{10373080} & 1T+2R & 5 & 2 & rigid & yes \\
      Parallel Gripper & 0 & 1 & 1  & rigid  & yes \\
      \textbf{Ours} & 1T+2R & 4/5 & 2  & soft & yes\\
      \bottomrule
      \end{tabular}
    }
  \end{minipage}%
  \hfill
  \begin{minipage}[t]{0.38\linewidth}
    \centering
    \vspace{0pt} 
    
    \vspace*{2.8em} 
    
    \includegraphics[width=\linewidth]{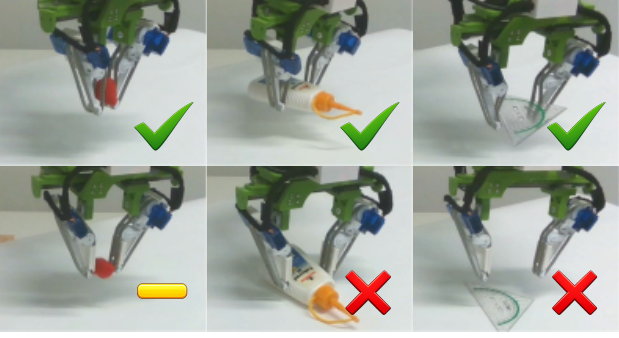}
    
    \captionof{figure}{\textbf{Comparison between the Belt-Finger (top) and a close variant (bottom) on picking and reorientation.}}
    \label{fig:compare-1}
  \end{minipage}
  
  \vspace*{-1.em}
\end{figure*}

\paragraph{Fingers with active surfaces}
Active surfaces enable continuous in-hand manipulation without complex multi-body kinematics \cite{jiang2025rotipbot}. For instance, two driven belts provide effective rolling and translational control \cite{spiers2018variable, tincani2013implementation, xiang2024adaptive, jiang2025rotipbot}. While incorporating rotatable fingertips further increases in-hand DoFs \cite{xu2024dtactive, yuan2020design}, it frequently results in overly complex \cite{xu2024dtactive} or over-actuated \cite{yuan2025tactile} systems.
\cref{tab:comprison-of-grippers} compares design complexity and manipulation ability across various grippers. We define "in-hand DoFs" as the available translational (T) and rotational (R) degrees of freedom, "kinematic chain" as the longest kinematic chain from the base to the contact points, and "portable finger" as the design's ease of reuse with other gripper bases.

Our work relates closely to the tri-finger belt \cite{10373080} and the rigid-support double-belt \cite{11082440}. Because rigid designs struggle with complex geometries \cite{11082440}, we contrast these approaches with our soft-belt design in \cref{sec:belt-design}.

\section{Hardware Design}
The hardware comprises three components: finger, belt, and gripper base.
\paragraph{Finger Design}
We enhance parallel grippers by actively shifting contact points to enable in-hand translation, roll, and pitch (\cref{fig:gripper-render}). 
The modular, plug-and-play fingers feature pointed tips and integrate three continuous-rotation servos with synchronous wheels (\cref{app:design-choices}).

\paragraph{Belt Design}\label{sec:belt-design}
Our primary hardware improvement is the \textbf{soft, deformable belt design}. Utilizing thin TPU95 belts \cite{TPU95A} with a high-friction coating and circular arc teeth, the system ensures stable wrapping and precise synchronization. Unlike prior rigid-support designs \cite{10373080} that simplify control but sacrifice dexterity (failing to grasp 3 of 5 curved objects in our tests, whereas ours reliably manipulated all 5, \cref{fig:compare-1}), our structure is distinct:
\begin{itemize}[nosep, leftmargin=2em]
    \item \textbf{Compliance \& Wrapping}: Soft belts passively deform for gentle, robust grasping of irregular surfaces.
    \item \textbf{Clutter Accessibility}: Narrow, pointed fingertips enable manipulation in cluttered spaces.
    \item \textbf{Low Cost}: Comprises 3D-printable and inexpensive off-the-shelf components.
    \item \textbf{Extensibility}: Internal volume permits future tactile sensor integration.
\end{itemize}
Omitting a rigid support introduces complex, non-constant sliding contacts. 
We solve this challenging control problem in the next chapter using our stochastic MPC (\cref{sec:MPC}) and a simple model. 
Furthermore, extensive benchmarking across teleoperation and modern VLA models (\cref{sec:vla}) demonstrates the system's practical viability and robustness for broader societal adoption.

\paragraph{Gripper Base \& Fabrication}
Our open-source, easily replicated base is mostly 3D-printed (FFF), bringing the total finger material cost to ${\sim}10$ euros (\cref{app:materials}). Driven by a Dynamixel MX28 motor with a plastic rack-and-pinion (100 mm range), it employs semi-impedance control:
\begin{equation}
 F_{\text{drive}} = K_p\cdot (p_{\mathrm{targ}}-p_{\mathrm{cur}}) + K_d \cdot (\dot{p}_{\mathrm{targ}}-\dot{p}_{\mathrm{cur}}) \label{eq:gripper_base_drive_force}
\end{equation}

\paragraph{Control}
Instead of raw motor velocities, we directly command theoretical object translation, roll, and pitch ($a$). Belt velocities $v^t_{i}$ are computed using a scaling constant $\gamma$:
\begin{equation}
    \begin{bmatrix}v^t_{1}\\v^t_{3}\\v^t_{4}\end{bmatrix}
= \gamma
\left[\begin{array}{@{}rrr@{}}1 & 1 & 1\\ 1 & 1 & -1 \\ 1 & -1 & -1 \end{array}\right]
\cdot
\left[\begin{array}{@{}l@{}}a^t_{\mathrm{trans}} \\ \aroll^t \\ a^t_{\mathrm{pitch}}\end{array}\right] \label{eq:mpc-contact-vel}
\end{equation}
  \vspace{-1.em}

\begin{figure}[htbp]
  \centering
  \hfill
  \begin{minipage}{0.40\columnwidth}
    \centering
    \includegraphics[width=\linewidth]{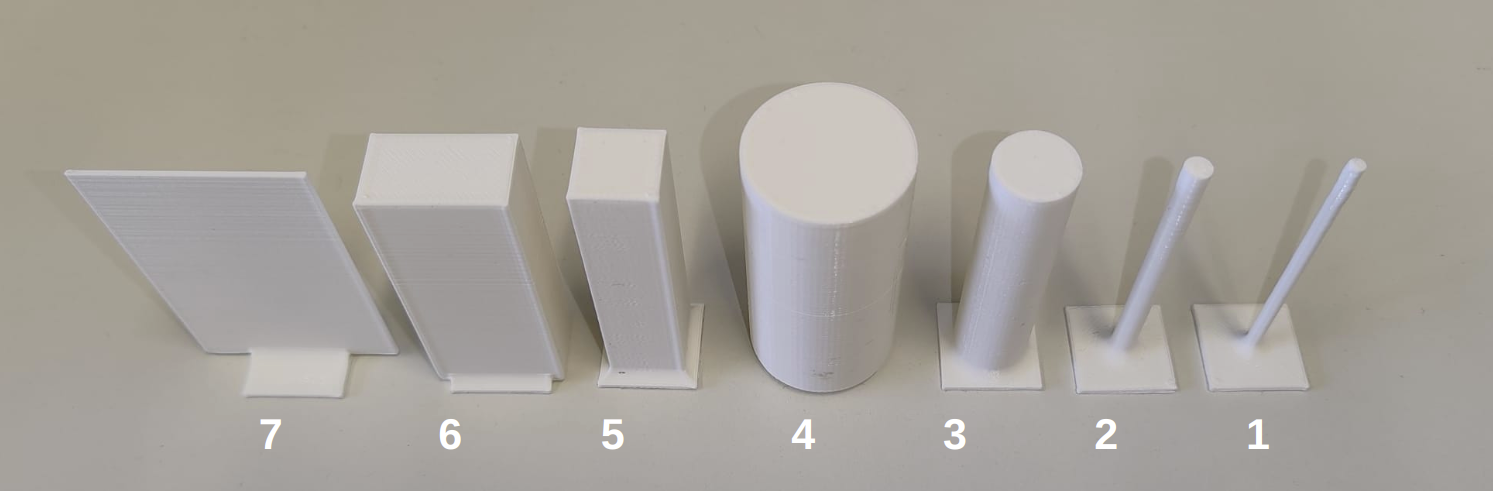}
    \label{fig:benchmarking-objects}
  \end{minipage}\hfill
  \begin{minipage}{0.55\columnwidth}
    \centering
    \includegraphics[width=\linewidth]{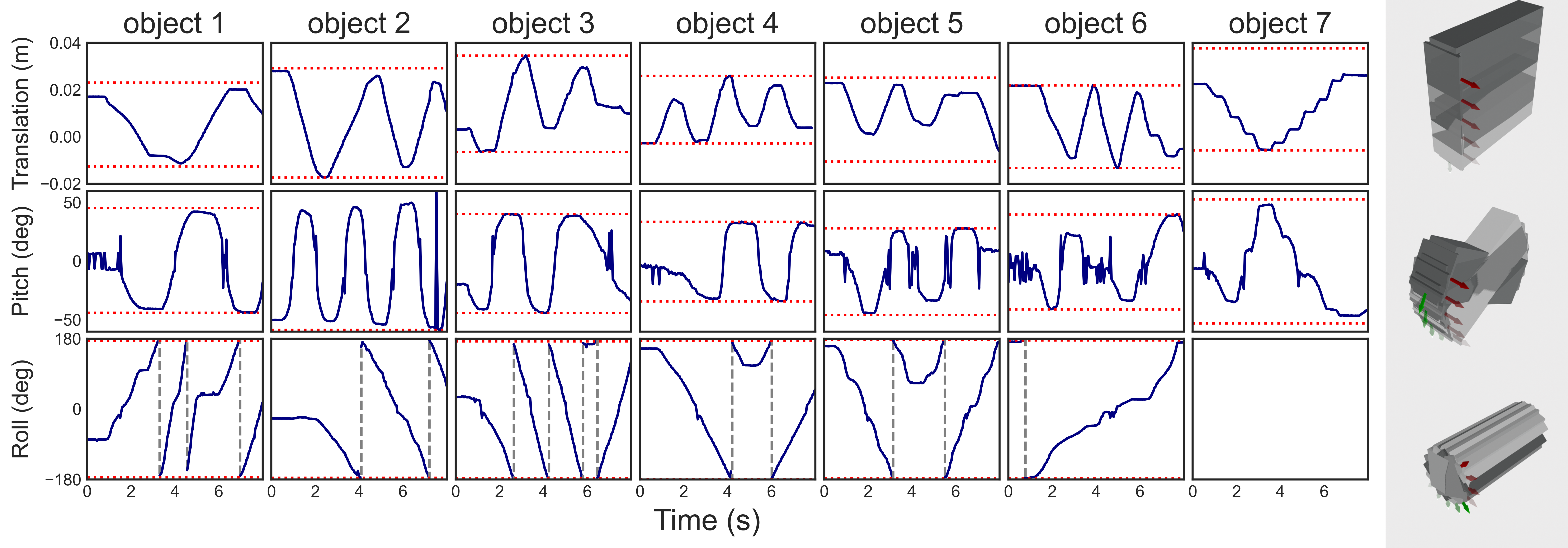}
    \label{fig:benchmarking-plot}
  \end{minipage}
  \vspace{-1.em}
    \caption{\textbf{In-hand motion benchmarking with teleoperation.} Left: AprilTag tracking for motion reconstruction. Right: Motion examples (plots of motion ($y$) versus time ($x$). Red dashed lines indicate max/min ranges. Columns represent objects, demonstrating smooth manipulability.}
  \vspace{-1.em}
\end{figure}

\section{Model Predictive Control (MPC)} \label{sec:MPC}

 To solve the challenging control problem introduced by the soft belts' non-constant sliding contacts, we propose an effective Model Predictive Control (MPC) framework for manipulating known object shapes. 
 The pipeline consists of three components: \textbf{a belt finger model, a zero-order optimizer, and a gripper opening estimator}. Manipulation performance is evaluated across three benchmarking objects of varying difficulty.

\subsection{Belt Finger Model} \label{sec:model}
The gripper's mixed rigid, deformable, and driven structure makes exact dynamic modeling challenging. However, MPC only requires a local, short-term approximation. We simplify our model using the following assumptions:
\begin{itemize}[nosep]
\item Contacts are sticky (no slip occurs).
\item Belts are approximated by their centerlines.
\item Belt motors supply sufficient power for the desired object motions.
\item Contacts are treated as points rather than surfaces.
\end{itemize}

\begin{figure}
\centering
\begin{minipage}[t]{.95\linewidth}
  \includegraphics[width=\linewidth]{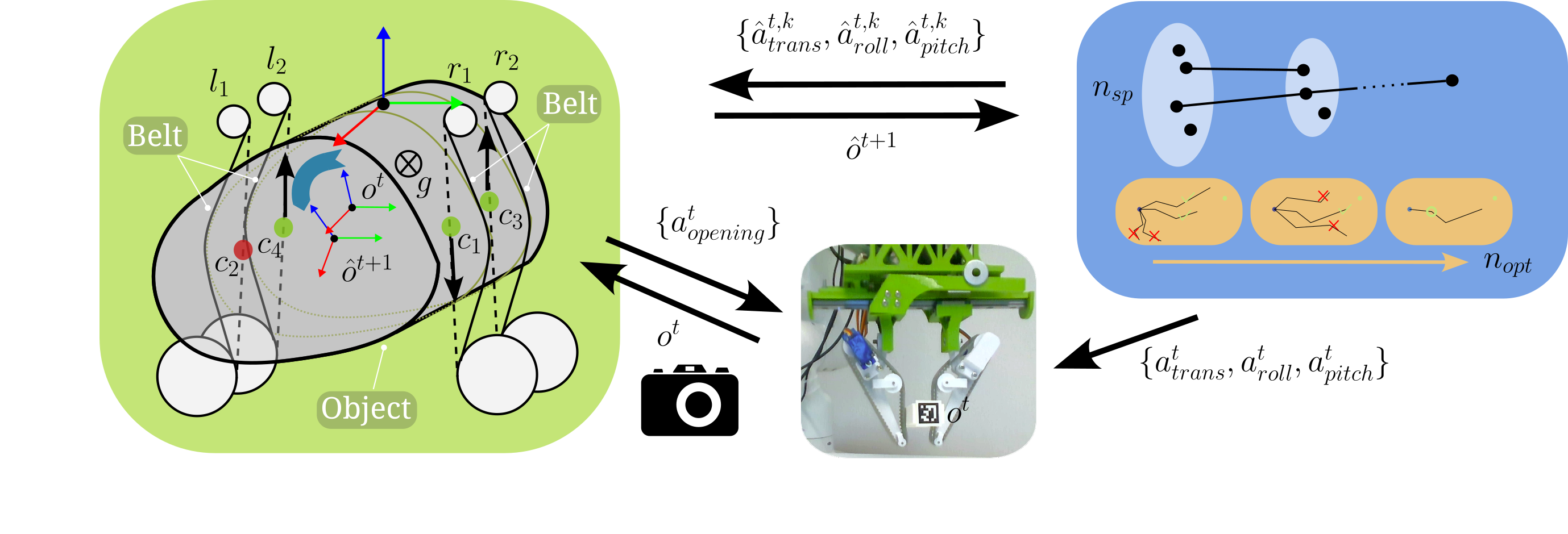}
  \vspace*{-1.5em}
  \caption{\textbf{Model predictive control setup.} The gripper model is shown in green, and the iCEM optimizer is shown in blue. $l_1,l_2$ and $r_1,r_2$ are left and right belts. $C=c_{1-4}$ are approximated contact points, where green means active controlled, while red is passive. In iCEM, for each executed action, $n_{opt}$ iterations of optimization will be performed, starting with $n_{sp}$ samples. During each optimization iteration, trajectories with $h$ steps will be sampled and evaluated by the sum of weighted distances. The sampling distribution for the next iteration is updated based on the top-$k$ highest-scoring samples from the previous iteration.}
  \label{fig:mpc}
  \vspace*{-1.em}
\end{minipage}
\end{figure}

As outlined in \cref{fig:mpc}, the model $M$ predicts the next object pose $\hat{o}^{t+1}$ based on the current pose $o^t$, object mesh $m$, and action $a^t = \{a^t_{\mathrm{trans}},a^t_{\mathrm{roll}},a^t_{\mathrm{pitch}}\}$ (excluding gripper opening). For each belt, the contact point $c^t_i$ is approximated by averaging all mesh-belt intersections (if they exist). Under the no-slip assumption (\cref{eq:mpc-contact-vel}), contact velocity $v^t_{c_i}$ equals the belt velocity $v^t_i$. Using standard rigid body dynamics (\cref{app:mpc:model}), we compute the object's rotational ($\omega^t_o$) and translational ($v^t_o$) velocities to derive $\hat{o}^{t+1}$. As shown in \cref{sec:ex2}, this simplified model effectively controls the gripper.

\subsection{Zero-order Optimizer}
Because the ray-tracing process renders our model non-differentiable, we employ the iCEM zeroth-order planner \cite{PinneriEtAl2020:iCEM} (\cref{fig:mpc}, bottom). iCEM iteratively refines a normal distribution of low-cost action sequences.
Given a target pose $o^{\mathrm{targ}}$, we sample $n_{\mathrm{sp}}$ time-correlated sequences over horizon $h$. Our model estimates the resulting trajectories $\{\hat{o}^{t+1},...,\hat{o}^{t+h}\}_{\times n_{\mathrm{sp}}}$. We evaluate each trajectory using a weighted distance cost function:
\vspace{-.5em}
\begin{equation}
\label{eq:mpc-cost-func}
J = \sum_{i}^{h} w_i\, \mathrm{dist}(\hat{o}^{t+i},o^{\mathrm{targ}})
\end{equation}
where $w_i=0.1^i$ prioritizes closer states, and $\mathrm{dist}(\hat{o}^{t+i}, o^{\mathrm{targ}}) = \sqrt{ \| \log_{SO(3)}(\hat{R}^{t+i}, R^{\mathrm{targ}}) \|_F^2 + \| t^{\mathrm{targ}} - \hat{t}^{t+i} \|_2^2 }$ serves as the canonical metric \cite{JMLR:v21:19-027}. Over $n_{\mathrm{opt}}$ iterations, the highest-scoring sequences guide the next sampling distribution. Only the first action $\hat{a}^{t+1}$ is executed before replanning from $o^{t+1}$.

\vspace*{-1em}
\begin{figure*}[!b]
\centering
\begin{subfigure}{.19\linewidth}
  \includegraphics[width=\linewidth]{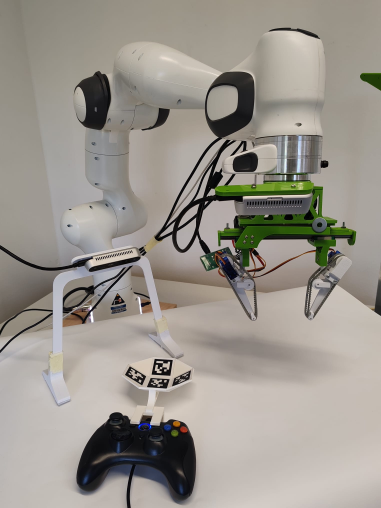}
\end{subfigure}%
\hspace{0.01\linewidth}
\begin{subfigure}{.604\linewidth}
  \includegraphics[width=\linewidth]{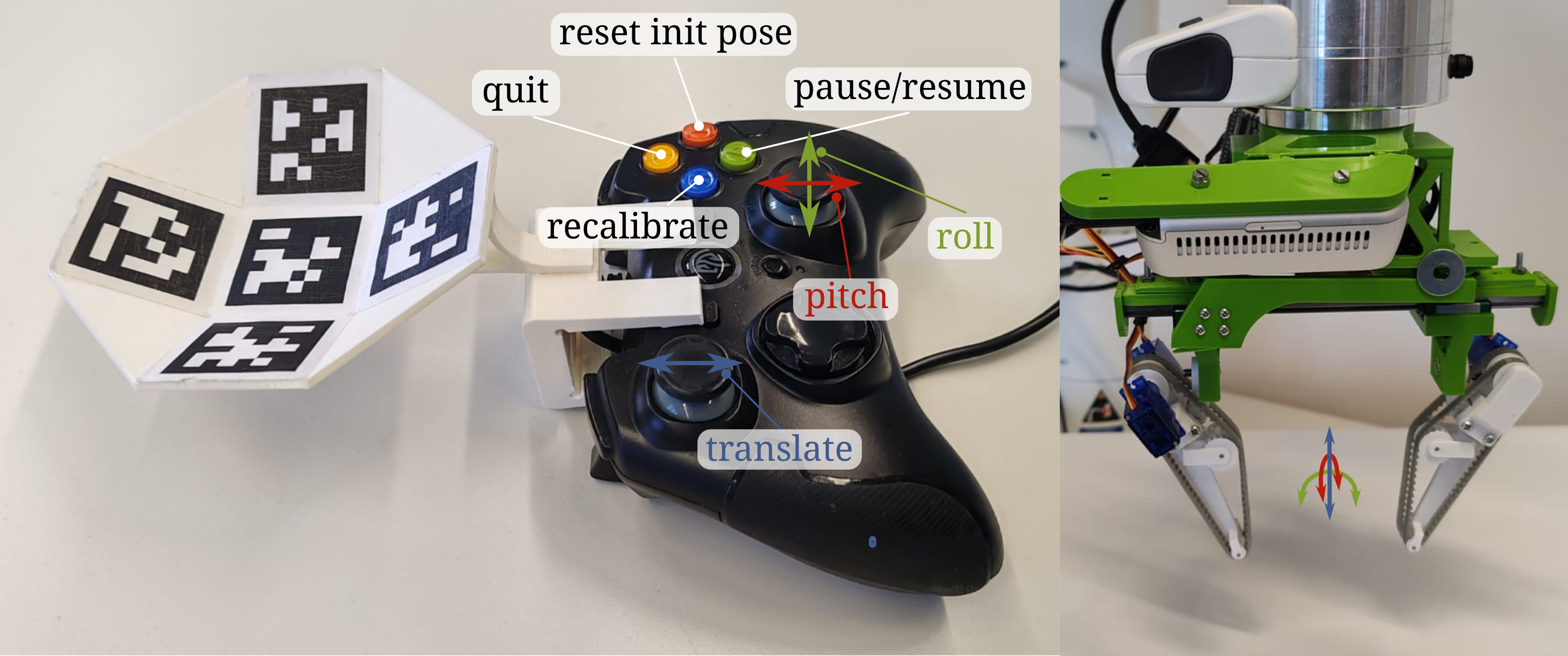}
\end{subfigure}%
  \caption{\textbf{The teleoperation setup for the \method{}} Left: with a 7-DoFs Franka Panda robot. Right: The motion mapping of the controller to the motion of the robot and the in-hand object (Middle).}
  \label{fig:teleoperation}
\end{figure*}

\subsection{Gripper Opening Estimator}
Grasp stability relies heavily on the gripper's opening width, $a_{\mathrm{opening}}$. Optimizing $a_{\mathrm{opening}}$ alongside spatial actions within iCEM is inefficient due to strict stability margins between the belts. To minimize required samples, we compute it via a feedforward prediction and an adaptive compensation term:
\begin{equation}
\label{eq:mpc-opening}
a_{\mathrm{opening}}^{t} = l_{b}(m, \hat{o}^{t+1}) + l^{t}_{\delta} 
\end{equation} 

\begin{wrapfigure}{R}{0.55\textwidth}
  \vspace{-15pt} 
  \begin{minipage}{\linewidth}
    \begin{algorithm}[H]
\footnotesize
\caption{In-Hand Manipulation with iCEM-based MPC}
\label{alg:mpc}
\begin{algorithmic}[1]  

\Require $m$, $o^{\mathrm{targ}}$, $o^{0}$

\State $l^{0}_{\delta} = 0$, $\hat{o}^0=o^0$
\For{$t = 0$ to $T$}\\
    \Comment{Plan for next action}
    \State $\{a^t_{\mathrm{trans}},a^t_{\mathrm{roll}},a^t_{\mathrm{pitch}}\},\hat{\C}^t,\hat{o}^{t+1}=$ \textit{iCEM}($m$,$o^t$,$o^{\mathrm{targ}}$)
    \State $\aopening = l_{b}(m, \hat{o}^{t+1}) + l^{t}_{\delta}$
    \State $o^{t+1} = \mathrm{execute}(\aopening,a^{t}_{\mathrm{trans}},a^{t}_{\mathrm{roll}},a^{t}_{\mathrm{pitch}})$

    \State distances = []\Comment{To find most plausible contact set}
    \For { \C{} in $\mathbb P(\hat{\C}^{t})$ } \Comment{Iterate through powerset}
       \State $\hat{o}_{\C} = \textit{M}(o^{t},a^{t}\mid m,\C)$ \Comment{Run model}
       \State $\mathrm{distances.append}((\C, \hat{o}_\C, \mathrm{dist}(o^{t+1},\hat{o}_{\C})))$
    \EndFor
    $\C^{t}_h,\hat{o}^{t+1}_{h}, \_ = \mathrm{minimum\_by\_dist(distances)}$
    \If{$|\C^{t}_h|<|\hat{\C}^{t}|$}
        \State $l^{t+1}_{\delta} = l^{t}_{\delta} - \Delta l$ \Comment{Under-pressure}
    \ElsIf{$\textit{dist}(o^{t+1},o^{t})<\textit{dist}(o^{t+1},\hat{o}^{t+1}_{h})$}
        \State $l^{t+1}_{\delta} = l^{t}_{\delta} + \Delta l$ \Comment{Over-pressure}
    \Else
        \State $l^{t+1}_{\delta} = l^{t}_{\delta}$
    \EndIf
\EndFor

\State \Return
\end{algorithmic}
\end{algorithm}
\end{minipage}
\vspace{-4em} 
\end{wrapfigure}
The forward term $l_{b}$ is the object's average segment length between the fingers at the predicted next pose. The adaptive term $l^{t}_{\delta}$ compensates for modeling errors, causing either under-pressure (gripper too open, lacking friction) or over-pressure (gripper too tight, restricting object movement).

To update $l_{\delta}$ (\cref{alg:mpc}), we extract the expected contact set $\hat\C^{t}$ from planning. After observing the true pose $o^{t+1}$, we evaluate all contact subsets to find the hindsight set $\C_h^t$ and predicted pose $\hat o^{t+1}_h$ that best explain the outcome. Under-pressure occurs if active contacts are fewer than predicted ($|\C^{t}_h| < |\hat{\C}^{t}|$). Over-pressure occurs if the object moves less than the model's minimum prediction ($\mathrm{dist}(o^t,o^{t+1}) < \mathrm{dist}(o^{t+1},\hat{o}^{t+1}_{h})$). In either scenario, $l_{\delta}$ is adjusted by a fixed step size $\Delta{l}$ (see \cref{app:mpc:pressure}).

\section{Teleoperation}

Controlling a 6-DoF robotic arm alongside our 4-DoF gripper presents an intuitive mapping challenge. 
Finding gestures and leader-follower methods unsuitable, we propose a simple teleoperation system using a controller and an external camera (\cref{fig:teleoperation}). 
The camera tracks the controller's marker array for global pose estimation. For the arm, we employ Cartesian impedance control \cite{zhu2022viola}, calculating the target pose as $p_{\mathrm{targ}} = p_{\Delta}\cdot p_{\mathrm{cur}}$. 
The controller's relative pose maps to the pose difference $p_{\Delta}$ via a second-order function, enabling fine-grained control near zero.
Crucially, utilizing this delta pose effectively yields significantly smoother trajectories compared to direct global pose mapping (e.g., VR tracking).
For the gripper, the right joystick dictates in-hand object rotation and pitch, the left joystick's $y$-axis controls translation, and the left trigger adjusts the opening size. This intuitive setup enabled an operator to rapidly adapt and record 750 trajectories in just 8 hours.


\section{Experiments}
Five experiments are conducted to address the following questions:
1.~ Do belt movements translate into three independent DoFs for the object? 
2.~How well does the \method{} work with MPC to manipulate objects in-hand?
3.~Does \method{} empower versatile in-hand manipulation of everyday objects?
4.~Can the VLA models be effectively finetuned on the gripper for complex tasks?
5.~What are the advantages of the extra in-hand DoFs compared to conventional fingers?

\subsection{Motion benchmarking for in-hand movement}

We benchmarked in-hand translation, pitch, and roll via \textbf{teleoperation} on seven diverse objects (\cref{fig:benchmarking-objects}, \cref{app:section-of-benchmarking-objects}). 
AprilTag \cite{olson2011apriltag} tracking recorded the motion trajectories (\cref{fig:benchmarking-plot}). 
\method{} successfully manipulated all objects except horizontally gripped object 7, achieving ranges of approximately 45 mm (translation), 85° (pitch), and continuous roll. 
We observed minor variances in speed and precision due to teleoperation, as well as slight pitch jitter caused by the restricted camera view.

\begin{figure*} 
  \centering
  
  \begin{minipage}[t]{0.52\linewidth}
    \centering
    \captionof{table}{\textsc{Measured payloads of test objects}} 
    \label{tab:payload-result}
    \vspace{0.2em}
    
    \adjustbox{max width=\linewidth}{
      \begin{tabular}{@{}c|c|c|c|c}
      \toprule
      \multirow{2}{*}{Object} &  \multirow{2}{*}{Size} & Translation & Roll          & Pitch       \\
                              &  & ($\mathrm{N}$)      & ($\mathrm{N\cdot mm}$) & ($\mathrm{N\cdot mm}$) \\
      \midrule
      \multirow{2}{*}{cuboid} & W*H*L & $-9.19$ & $-99.93$ & $-51.11$\\
                              &    20*15*60    & $+2.92$ & $+73.37$ & $+30.72$ \\\cline{1-5}
      \multirow{2}{*}{cylinder} & D*L & $-6.8$ & $-65.89$ & $-37.22$ \\
                                &   20*60    & $+1.86$ & $+49.31$ & $+24.93$ \\ \cline{1-5}
      \multirow{2}{*}{cone} & d*D*L & $-7.86$ & $-75.4$ & $-37.24$  \\
                            &     18*24*60   &$+2.87$ & $+59.14$ & $+24.27$\\ \cline{1-5}
      \multirow{2}{*}{sphere} & D & $-9.24$ & $-193.07$ & $-56.25$ \\
                              &    40   & $+4.32$ & $+137.81$ & $+36.39$\\
      \bottomrule
      \end{tabular}
    }
  \end{minipage}
  \hfill 
  \begin{minipage}[t]{0.45\linewidth}
    \centering
    \vspace{0pt} 
    \includegraphics[width=\linewidth]{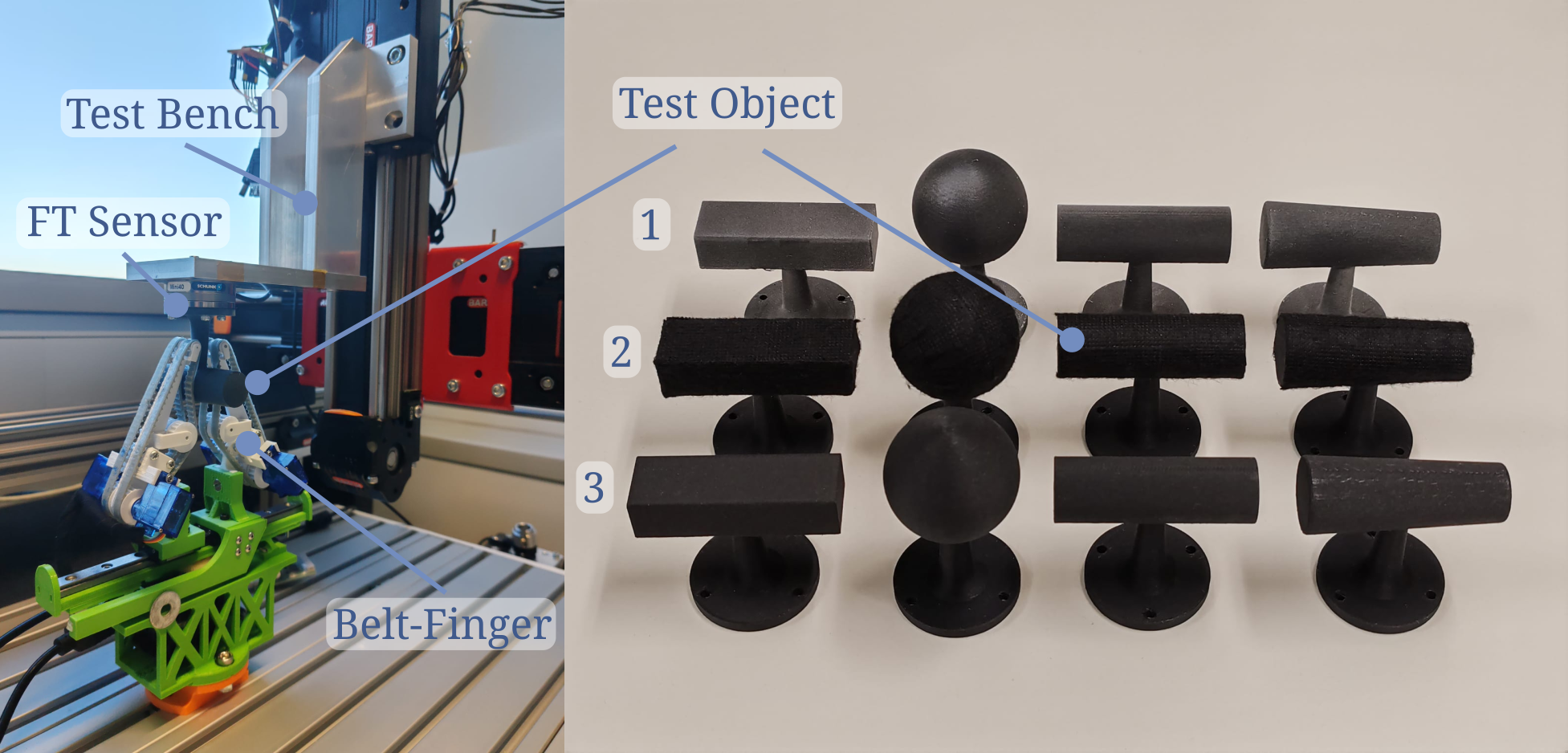}
    \captionof{figure}{\textbf{Payload benchmarking setup.} The in-hand payloads of four shapes are measured. The three surface material are listed in \cref{tab:material-payload-objects}.}
    \label{fig:payload-ex-setup}
  \end{minipage}
\vspace*{-1em}
\end{figure*}

Maximum force-torques were measured across four objects with three surface materials (\cref{fig:payload-ex-setup,tab:material-payload-objects,tab:payload-result,app:payload-result-with-std}). 
\method{} achieved maximums of 9 N (translation), 100 N·mm (roll), and 50 N·mm (pitch). 
Non-parallel finger design caused asymmetric translational forces (\cref{app:design-choices}). 
Despite a 7.5\% payload decay after moderate use (\cref{app:friction}), \method{} effectively handles most daily objects under 400 g.
Further benchmarking details are in \cref{app:specs}.

\subsection{Object manipulation through the iCEM-based MPC} \label{sec:ex2}

\begin{figure}[htbp]
\centering
\begin{subfigure}[t]{.48\linewidth}  
  \centering
  \includegraphics[width=\linewidth]{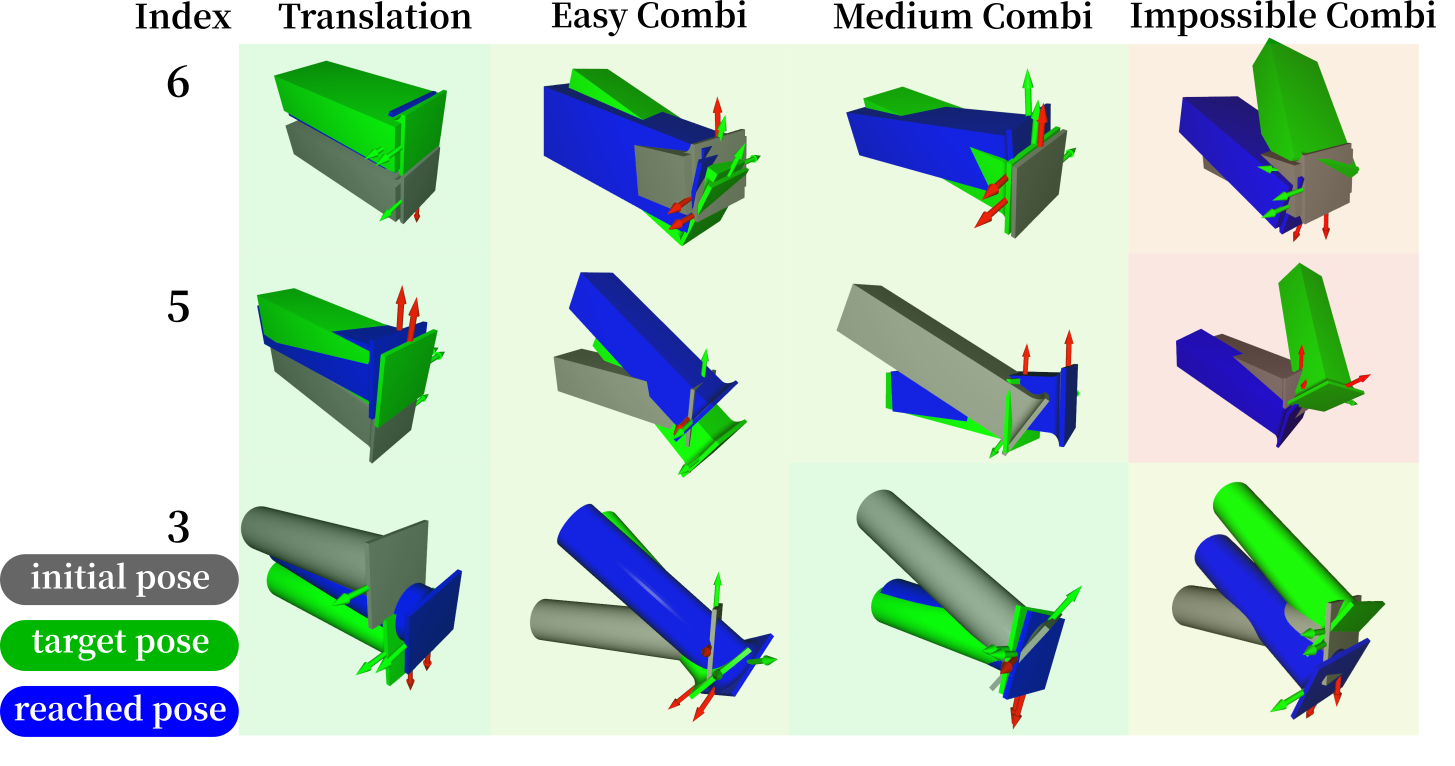}
  \caption{\textbf{Initial, target, and reached poses...} Both \emph{Easy Combi} and \emph{Medium Combi} require in-hand rotation and translation.}
  \label{fig:mpc-ex}
\end{subfigure}\hfill 
\begin{subfigure}[t]{.48\linewidth}
  \centering
  \includegraphics[width=\linewidth]{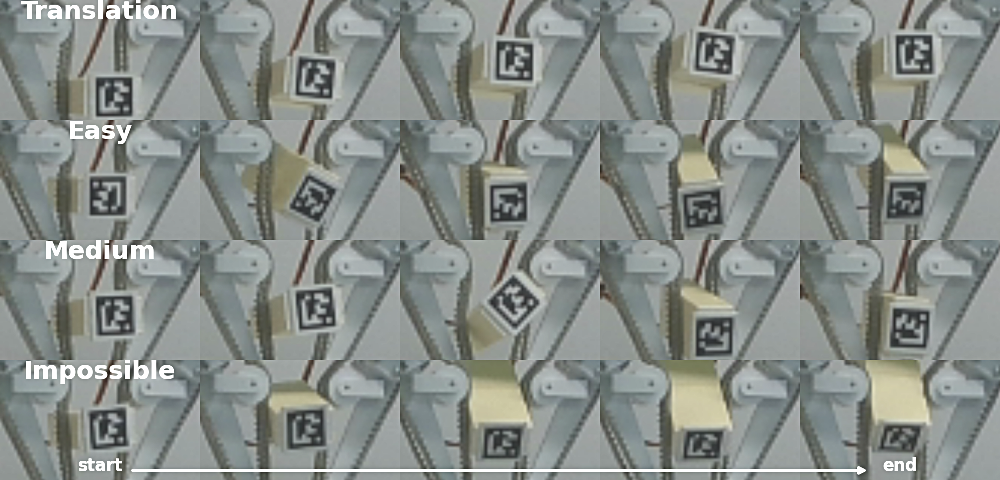}
  \caption{\textbf{The MPC rollout from object 6.} Time steps are downsampled to 5.}
  \label{fig:mpc-ex-real}
\end{subfigure}
\caption{Comparison of MPC rollouts in simulation and real-world environments.}
\label{fig:combined-mpc}
\end{figure}

\begin{wraptable}{r}{0.45\textwidth}
  \centering
  \caption{\textsc{The final pose error of each target difficulty}} 
  \label{tab:pose-error-mpc}
  \vspace{.2em}
  
  \adjustbox{max width=\linewidth}{
    \begin{tabular}{@{}c|c|c@{}}
      \toprule
      \textbf{Target} & \textbf{Position Error} & \textbf{Orientation Error} \\
      \midrule
      Translation & $4.8\pm 1.5\text{mm}$ &  $9.1 \pm 3.3 ^\circ$ \\
      Easy Combi & $8.1\pm3.7\text{mm}$ & $20.0 \pm 10.5 ^\circ$ \\
      Medium Combi & $6.7\pm4.1\text{mm}$ & $12.1 \pm 7.3 ^\circ$ \\
      Impossible Combi & $10.9\pm4.5\text{mm}$ & $42.1 \pm 7.0 ^\circ$ \\
      \bottomrule
    \end{tabular}
  }
  \vspace{-1em} 
\end{wraptable}
We evaluate the proposed \textbf{MPC} on three benchmark objects (indices 3, 5, and 6 in \cref{tab:section-of-benchmarking-objects}).
With a stationary robotic arm, the gripper manipulates objects toward target poses tracked via AprilTag. 
Targets span four difficulties: translation-only, easy, medium, and impossible combinations (roll, pitch, and translation). 
Control parameters are set as $n_{\mathrm{sp}}=50$, $n_{\mathrm{opt}}=3$, $h=2$, $\Delta t=0.2\,\mathrm{s}$, $\Delta l=3\,\mathrm{mm}$, and $w=[1.0,0.1]$. 
The initial, target, and end poses are visualized in \cref{fig:mpc-ex,fig:mpc-ex-real}.


Results demonstrate the \textbf{MPC} effectively reaches targets in easy and medium cases (the loss is notably optimized against orientation). 
The \textbf{MPC} minimizes errors even for theoretically unreachable targets in the combi-cases which violate the force-closure.
Residual errors (\cref{tab:pose-error-mpc}) can be compensated by the arm in operational space. 
Further tests on five complex objects \cite{mahler2017dex} (\cref{app:complex-object-manipulation}) yielded average pose errors of: $9.4 \pm 5\,\text{mm}$, $14 \pm 16^\circ$ \textbf{(translation)}; $14 \pm 4\,\text{mm}$, $16 \pm 23^\circ$ \textbf{(roll)}; $13 \pm 6\,\text{mm}$, $13 \pm 11^\circ$ \textbf{(pitch)}; and $13 \pm 8\,\text{mm}$, $33 \pm 25^\circ$ \textbf{(combi)}.

\subsection{Manipulation of daily objects}\label{sec:ex3}

We demonstrate the gripper's teleoperated manipulation capabilities through two challenging tasks.

\begin{figure}
  \centering
  
  \begin{minipage}[t]{0.48\linewidth}
    \centering
    \vspace{0pt} 
    \includegraphics[width=\linewidth]{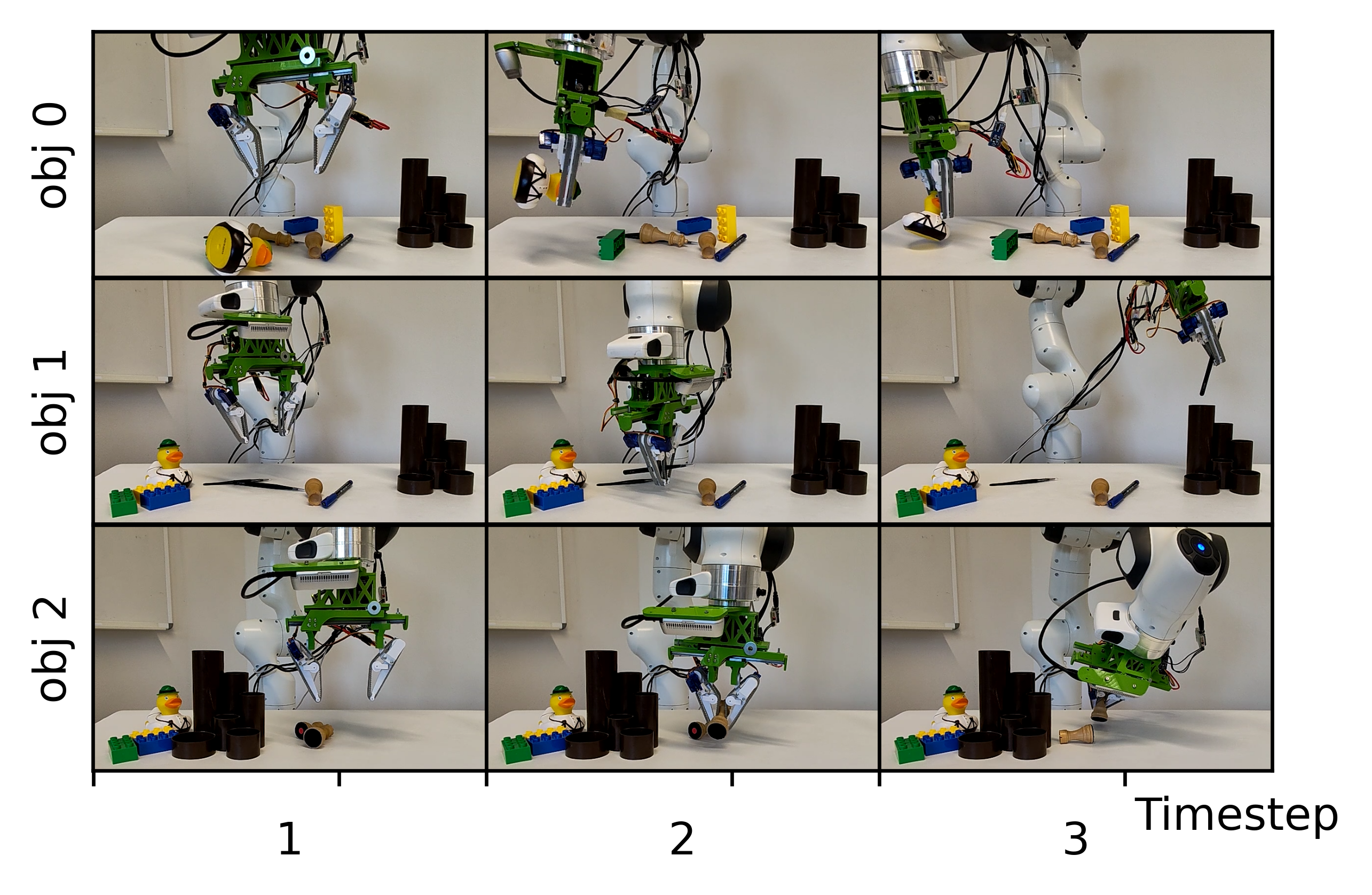}
    \vspace*{-1em}
    \caption{\textbf{Teleoperated \method.} The operator teleoperates the robotic arm and the gripper for a table cleaning task that involves complex in-hand motion. }
    \label{fig:table-clean}
  \end{minipage}%
  \hfill 
  \begin{minipage}[t]{0.48\linewidth}
    \centering
    \vspace{0pt} 
    \includegraphics[width=\linewidth]{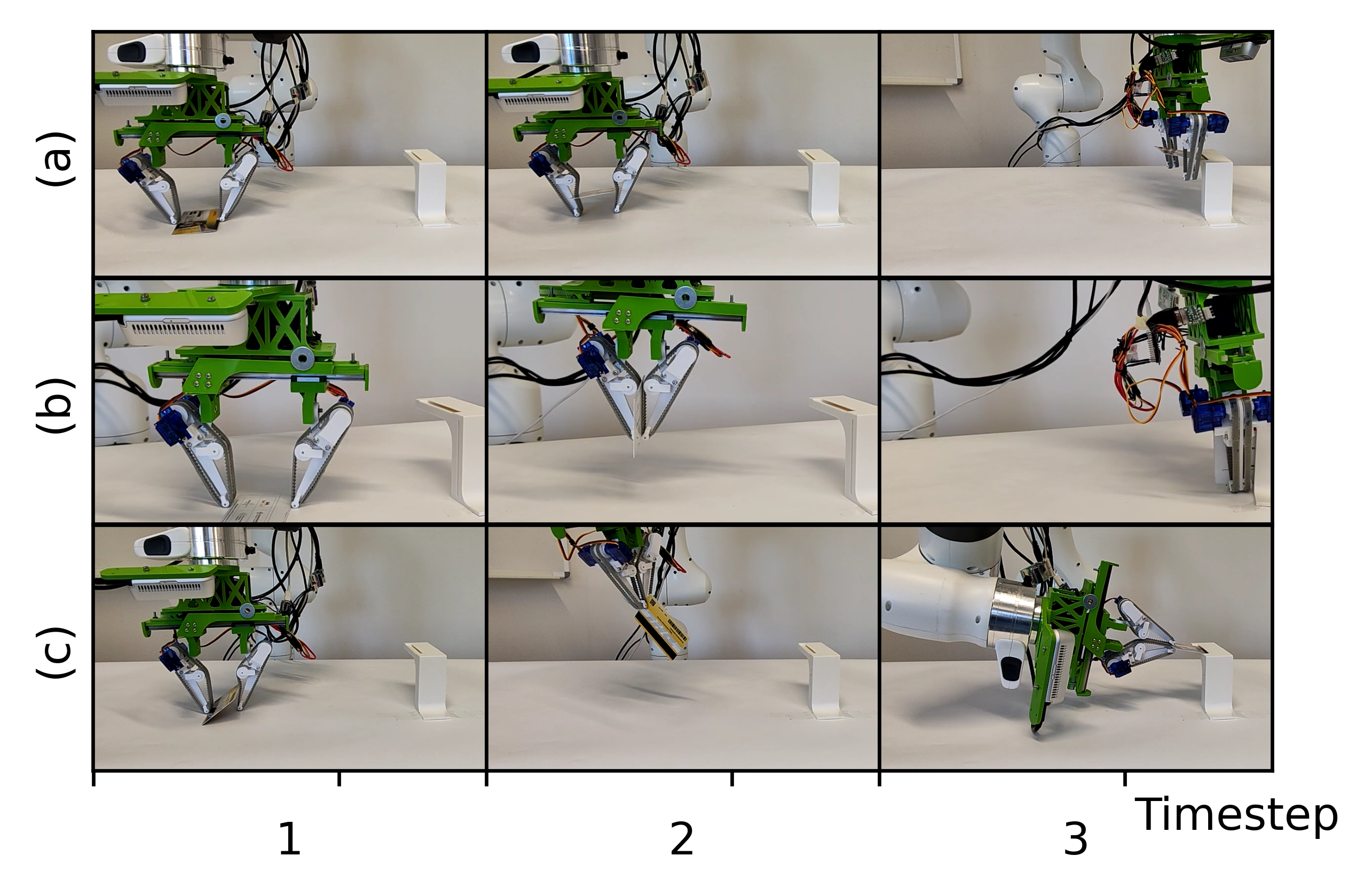}
    \vspace*{-1em}
    \caption{\textbf{Showcasing a card insertion task using the teleoperation setup.} The card is picked up from the table and inserted horizontally in both (a) and (c), using different strategies. }
    \label{fig:card-insertion}
  \end{minipage}
    \vspace*{-1.5em}
\end{figure}

\textbf{Task One: Table Cleaning.} The operator picks, reorients, and neatly places irregularly shaped objects in constrained spaces—a task difficult for standard parallel grippers. As shown in \cref{fig:table-clean}, the gripper efficiently flips LEGOs, rapidly reorients pens and tweezers using combined translational and pitch motions, and powerfully moves heavy objects without requiring large robotic arm movements.

\textbf{Task Two: Card Insertion.} The operator picks a bank card and inserts it into vertical and $15^{\circ}$ sloped slots. \cref{fig:card-insertion} shows the gripper picking the card vertically and horizontally. The fingers' fine-grained pitch motion enables precise alignment, minimizing the necessary arm adjustments.



\subsection{Object manipulation through imitation learning (IL) }\label{sec:vla} 

\begin{figure}
\centering
  \vspace*{-1.em}
  \includegraphics[width=\linewidth]{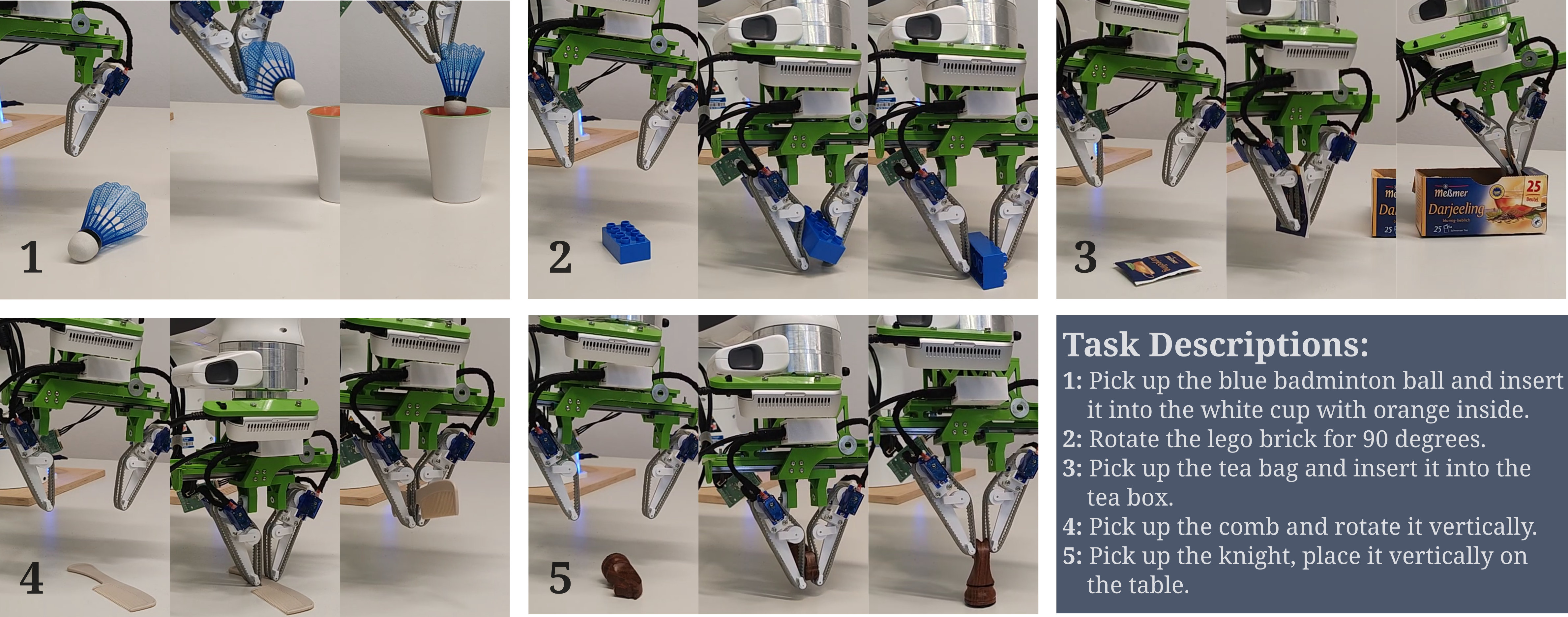}
\caption{\textbf{VLA control of the \method{}.} We evaluate five challenging daily tasks requiring complex in-hand manipulation. Finetuned VLA models successfully leverage the soft belts to achieve high success rates, demonstrating closed-loop corrections on tasks inaccessible to standard parallel grippers.}
  \label{fig:imitation}
  \vspace*{-1.5em}
\end{figure}

 
\begin{wraptable}{r}{0.5\textwidth} 
\centering
\caption{\textsc{Success rate for finituned VLA}} \label{tab:vla-success-rate}
\vspace{.8em}
\adjustbox{max width=\linewidth}{
\begin{tabular}{@{}c|c|c|c@{}}
\toprule
\textbf{Tasks} & \textbf{ACT} & \textbf{$\pi_{0.5}$} & \textbf{GR00T N1.7} \\
\midrule
badminton\_in\_cup & 60\% & 80\% & 70\%\\
lego\_reorientation & 80\% & 95\% & 100\%\\
pick\_and\_insert\_tea\_bag & 0\% & 85\% & 85\%\\
serve\_comb & 60\% & 70\% & 60\%\\
recover\_knight & 0\% & 60\% & 75\%\\
\bottomrule
\end{tabular}
\vspace{-1.em}
}
\end{wraptable}
While large robotic datasets for Vision-Language-Action (VLA) models exist \cite{o2024open}, they predominantly rely on standard two-finger grippers or human hands. Adapting pre-trained VLAs to a novel multi-DoF mechanism, such as \method{}, is an unprecedented and non-trivial challenge.
To pioneer this integration and demonstrate our design's potential to fundamentally transform generalist robotics, we evaluated the \method{} on five daily tasks (\cref{fig:imitation}). Crucially, these tasks require complex in-hand manipulation that is highly prohibitive for conventional parallel grippers. Using our intuitive teleoperation setup, we rapidly collected 150 diverse trajectories per task in just 8 hours, expanding the robotic system's action space to 10 dimensions.

We fine-tuned three state-of-the-art models—$\pi_{0.5}$ \cite{black2025pi_}, GR00T N1.7 \cite{gr00tn1_2025}, and ACT \cite{zhao2023learning}—and report the success rates across 20 rollouts in \cref{tab:vla-success-rate} (training and task details in \cref{supp:vla}). 
The results conclusively demonstrate that modern VLAs can successfully leverage the \method{} for complex in-hand manipulation. Notably, GR00T N1.7 and $\pi_{0.5}$ achieved exceptionally high success rates. We observed emergent, closed-loop correction behaviors of the policies using the active belts during inference. This seamless VLA-compatibility highlights the gripper's readiness for widespread adoption to tackle real-world, high-dexterity challenges.


\subsection{Benchmarking for manipulation between w/- or w/o in-hand motion}



\begin{wrapfigure}{R}{0.4\textwidth}
  \vspace{-1.5em} 
  \centering
  \includegraphics[width=\linewidth]{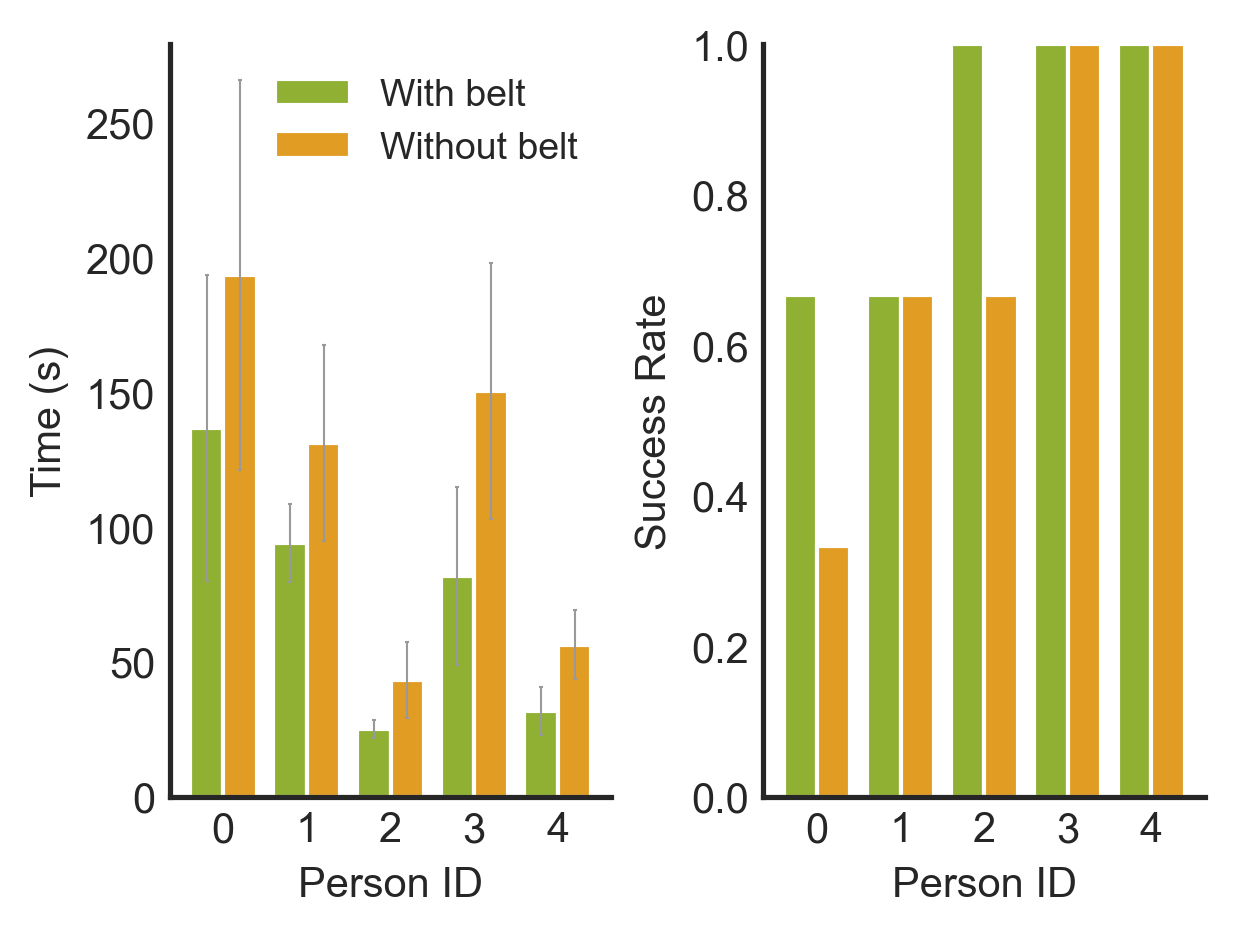}
  \vspace{-1em} 
  \caption{\textbf{User study.} The success rate and total time used for finishing the pen insertion task across participants.}
  \label{fig:pen-insertion}
  \vspace{-2em} 
\end{wrapfigure}

We evaluated the teleoperated pick-rotate-insert task (picking a randomly positioned pen and inserting it into a holder) comparing grippers with and without in-hand manipulation. Five participants completed three trials each. As shown in \cref{fig:pen-insertion}, using belt fingers significantly reduced both failure rates and execution times. High time variances among users stemmed from varying robotics experience affecting their spatial mapping abilities. Post-experiment, all participants preferred \method{}, citing its intuitive in-hand reorientation over complex, singularity-prone arm motions. Furthermore, a 10-task benchmark (\cref{app:benchmark}) demonstrated a 100\% success rate for \method{}, compared to a 50\% failure rate with conventional parallel fingers.

\vspace*{-0.5em}
\section{Limitations}
The \method{} has several limitations: (i) due to the double-belt layout the pitch for small objects is limited and the usable workspace for very thin/large/concave shapes is reduced; (ii) manipulatability depends on friction/adhesion of the coating and pressure distribution, making rotations nonlinear and less repeatable across surfaces, contamination, and wear; (iii) the TPU belts introduce elasticity/hysteresis that can reduce precision, and long-term wear (stretching/tooth wear/coating degradation) is not yet characterized, although we have not had problems; (iv) there are force/torque limits set by belt traction and compact actuators; and (v) the in-hand motion is conditioned on both robot dynamics and object properties, which are partially simplified in the MPC controller. 
\vspace*{-0.5em}

\section{CONCLUSIONS}


In this work, we introduce a low-cost \method{} gripper that significantly enhances the dexterous in-hand manipulation capabilities of conventional parallel two-finger grippers. To efficiently collect human demonstrations for challenging tasks, such as card insertion and table cleaning, we designed an intuitive and robust teleoperation system. Furthermore, we implemented an iCEM-based MPC to achieve precise, zero-shot in-hand manipulation. Crucially, our experimental evaluation of a fine-tuned Vision-Language-Action (VLA) model provides valuable insights into how such advanced learning mechanisms are exceptionally well-suited for modern robotic systems, bridging high-level reasoning with low-level dexterity. Overall, our evaluations demonstrate that this design outperforms classic parallel grippers in usability, efficiency, and user satisfaction.

Looking ahead, data-driven model-based reinforcement learning is a promising direction to address current modeling complexities. Another highly relevant direction is integrating additional sensing capabilities into the fingers to further empower these learning models.


\newpage

\bibliography{refs}

\newpage

\appendix

\section{Specification of the belt finger gripper}\label{app:specs}


To give an intuitive understanding of the in-hand manipulation capabilities of the \method{}, the maximum force and torques of each manipulation DoFs are measured with four types of objects on the test bench shown in \cref{fig:payload-ex-setup}. For each object, three different surface materials are chosen as shown in \cref{tab:material-payload-objects}.
\begin{wraptable}{r}{0.45\textwidth}
  \centering
  \vspace{-1em} 
  
  \caption{\textsc{Surface materials of the objects for force testing}} 
  \label{tab:material-payload-objects}
  
  \adjustbox{max width=\linewidth}{
    \begin{tabular}{@{}c|c|c@{}}
      \toprule
      Index & Material & Vendor \\
      \midrule
      1 & Non-Slip Coating & Odif\\
      2 & Polyester Fabric Tape & YC Group \\
      3 & PLA-CF  & Bambu Lab\\
      \bottomrule
    \end{tabular}
  }
  
  \vspace{-1em} 
\end{wraptable}
The test object is mounted under a 6-DoF force-torque sensor (ATI Mini40), which is mounted on the arm of the test bench with movable x-y-z axes. The gripper is fixed upwards on the test bench. During the measurement, the test object is first moved to the center of the \method{}. Then, the gripper is closed to a fixed opening width for each object. Then, the belt will rotate to perform the translation, rotation, and pitch movement from zero speed to positive and negative maximum speeds. The averaged force-torque values are shown in \cref{tab:payload-result}, where the belt deformation is the largest deformation after closing the gripper, and pull is the largest force detected during the pulling motion. Forces/torques from positive and negative directions are measured separately (top row: negative, bottom row: positive).

The translational force lies between ~6 and 9N, which further increases when the pressure between fingers and object is larger (sphere).

The maximum torque is nearly double for roll movement than for pitch movement.
A major reason is the large friction torque caused by the relative rotation of the object on the belt surface.
Objects that are lighter than 400g (e.g, Care Products, Office Supplies) can be mostly manipulated by the \method{}.
Heavier objects require careful handling to keep their center of mass aligned between the belts.

One concern is whether the belt can fit tapered surfaces, where each pair of belts may have different contact forces. The measurement of all three key factors stay similar level for both cylinder and cone, indicating strong adaptability.

High-voltage motors and lubrication can be adopted to increase the manipulation capabilities. The longer finger design also contributes to the wrapping positively.

\section{Payload test full result}\label{app:payload-result-with-std}
\begin{table}[H]

\centering
\caption{\textsc{Measured payloads of test objects}} \label{tab:payload-result-with-std}\vspace{.2em}
\vspace{-.2em}

For cells with a double row, the top row shows negative measurements for

translation, roll, and, pitch, while the bottom row shows positive ones.
\vspace{.5em}
\adjustbox{max width=\linewidth}{
\begin{tabular}{@{}c|c|c|c|c|c}
\toprule
\multirow{2}{*}{Object} & Size   & Opening & Translation & Roll          & Pitch       \\
                        & ($\mathrm{mm}$) & ($\mathrm{mm}$)  &  ($\mathrm{N}$)      & ($\mathrm{N\cdot mm}$) & ($\mathrm{N\cdot mm}$) \\

\midrule
\multirow{2}{*}{cuboid} & W*H*L & \multirow{2}{*}{10} &  $-9.19 \pm 2.68$ & $-99.93 \pm 17.85$ & $-51.11 \pm 7.94$\\
                        &    20*15*60    & & $+2.92 \pm 1.28$ & $+73.37 \pm 16.25$ & $+30.72 \pm 5.53$ \\\cline{2-6}
\multirow{2}{*}{cylinder} & D*L & \multirow{2}{*}{10} & $-6.8 \pm 1.15$ & $-65.89 \pm 6.95$ & $-37.22 \pm 4.35$ \\
                          &   20*60    & & $+1.86 \pm 0.41$ & $+49.31 \pm 4.25$ & $+24.93 \pm 2.6$ \\ \cline{2-6}
\multirow{2}{*}{cone} & d*D*L & \multirow{2}{*}{10} & $-7.86 \pm 1.46$ & $-75.4 \pm 7.84$ & $-37.24 \pm 4.61$  \\
                      &     18*24*60   & & $+2.87 \pm 0.57$ & $+59.14 \pm 6.43$ & $+24.27 \pm 3.05$\\ \cline{2-6}
\multirow{2}{*}{sphere} & D & \multirow{2}{*}{25} & $-9.24 \pm 1.16$ & $-193.07 \pm 9.97$ & $-56.25 \pm 4.74$ \\
                        &    40   & & $+4.32 \pm 0.88$ & $+137.81 \pm 11.19$ & $+36.39 \pm 6.3$\\
\bottomrule
\end{tabular}
}
\end{table}

\section{Materials}\label{app:materials}

\begin{table}[H]
\centering
\caption{\textsc{Components with their materials}} \label{tab:material}\vspace{-.2em}
\adjustbox{max width=\linewidth}{
\begin{tabular}{@{}l|c|c@{}}
\toprule
\textbf{Parts} & \textbf{Material} & \textbf{origin} \\
\midrule
Finger and gripper base, synchronous wheels & PLA & 3D printed \\
Synchronous belt & TPU 95 & 3D printed \\
Coating layer & nano tape & ordered \\
Transmission rack and pinion & igutek P360 & ordered\\
Linear track, axes, fastener & stainless steel & ordered\\
Servo motor & --- & ordered\\
\bottomrule
\end{tabular}
}
\end{table}

\section{Benchmarking Specifications} \label{app:section-of-benchmarking-objects}

\begin{table} [H]
\centering
\caption{\textsc{Section parameters of the benchmarking objects}} \label{tab:section-of-benchmarking-objects}\vspace{.2em}
\adjustbox{max width=\linewidth}{
\begin{tabular}{@{}c|c|c@{}}
\toprule
Index & Shape & Size \\
\midrule
1 & cylinder  & D: 3 mm\\
2 & cylinder  & D: 5 mm\\
3 & cylinder  & D: 15 mm\\
4 & cylinder  & D: 30 mm\\
5 & square    & W*H: 15*15 mm\\
6 & rectangle & W*H: 15*25 mm\\
7 & line (card)& W*H: 40*2 mm\\
\bottomrule
\end{tabular}
}
\end{table}

\section{Benchmarking between belt finger and parallel finger}\label{app:benchmark}
Ten daily tasks are performed with \method{} and a conventional parallel gripper using teleoperation.
The task descriptions are listed in \cref{tab:task-description}.

\begin{table}
\centering
\caption{\textsc{Descriptions for daily tasks}} \label{tab:task-description}\vspace{.2em}
\adjustbox{max width=\linewidth}{
\begin{tabular}{@{}c|l@{}}
\toprule
\textbf{Task Name} & \textbf{Description} \\
\midrule
Allen & pick up the allen key in a manipulatable way \\
Bottle & pick up and screw on the lid\\
Cloth & pick up the cloth\\
Cup & flip the cup\\
Knife & Pick up a knife for cutting.\\
Nail & pick up the nail.\\
Paper & pick up the paper without creasing it.\\
Pepper & right the pepper shaker\\
Plate & pick up the plate safely\\
USB & pick up and insert the USB into the hub\\
\bottomrule
\end{tabular}
}
\end{table}

\begin{figure*}
\begin{subfigure}{.49\linewidth}
  \includegraphics[width=\linewidth]{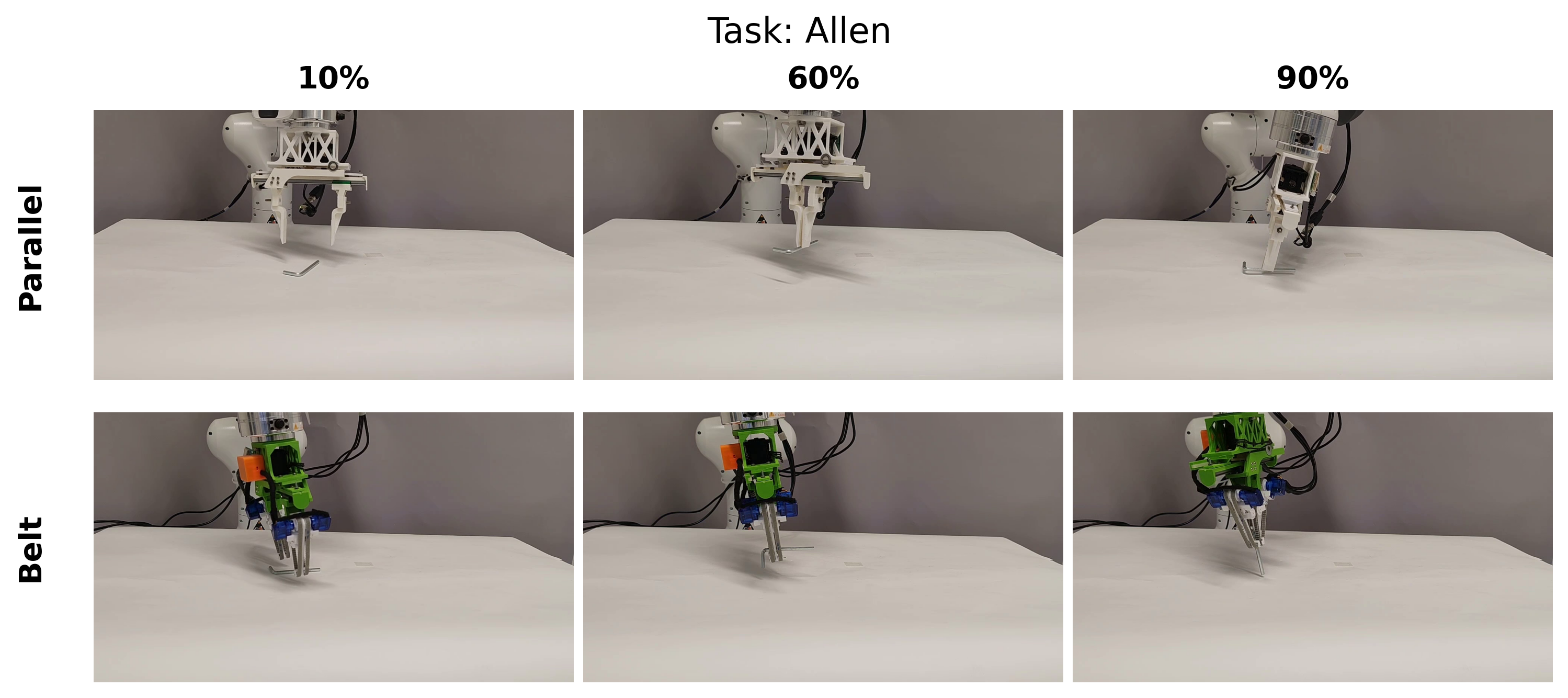}
\end{subfigure}
\begin{subfigure}{.49\linewidth}
  \includegraphics[width=\linewidth]{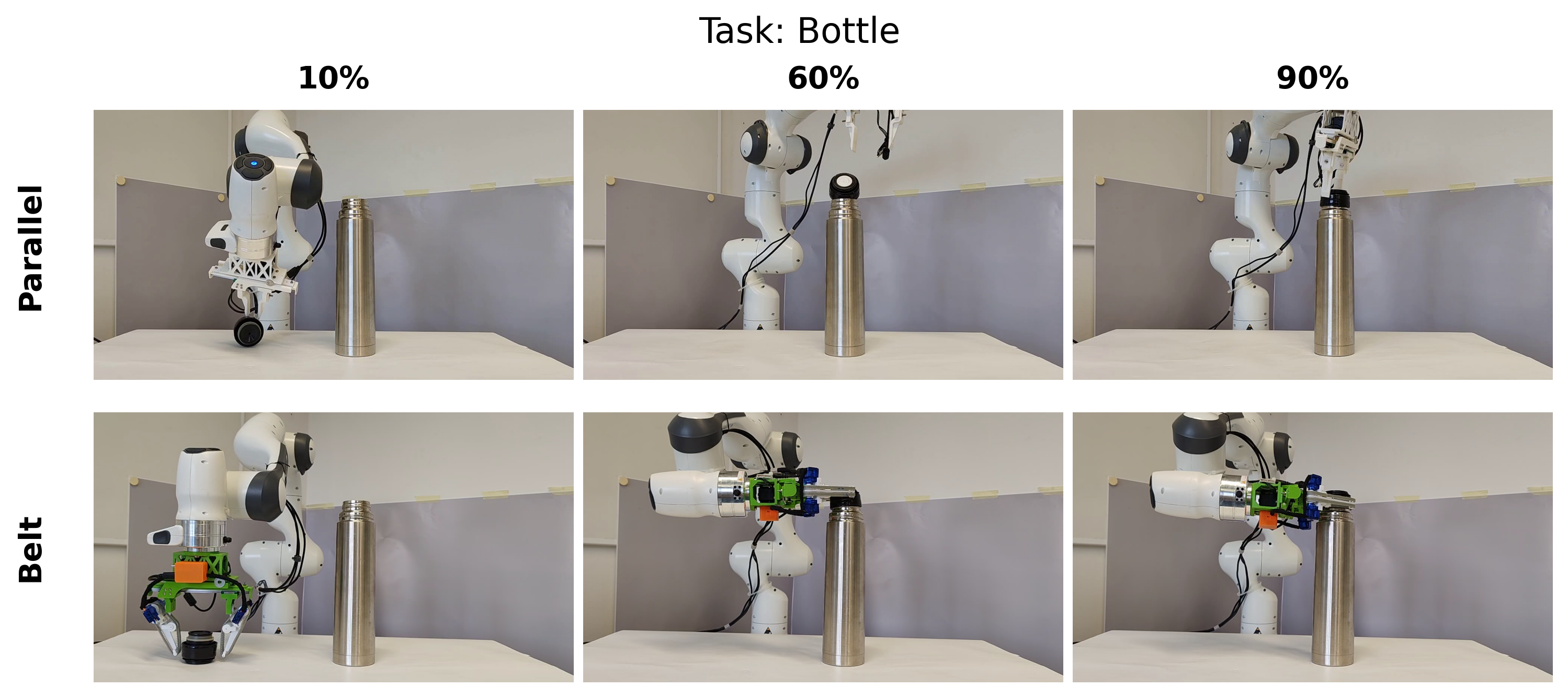}
\end{subfigure}
\begin{subfigure}{.49\linewidth}
  \includegraphics[width=\linewidth]{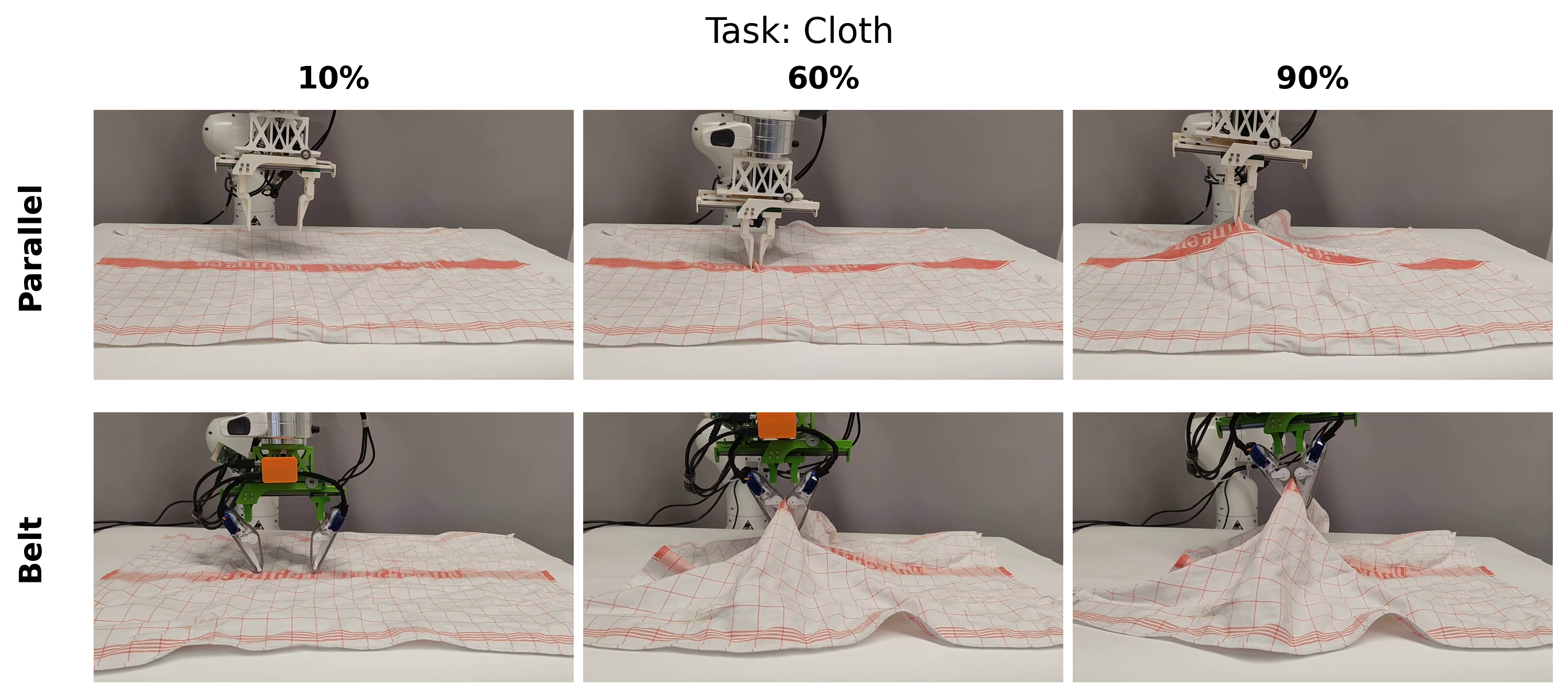}
\end{subfigure}
\begin{subfigure}{.49\linewidth}
  \includegraphics[width=\linewidth]{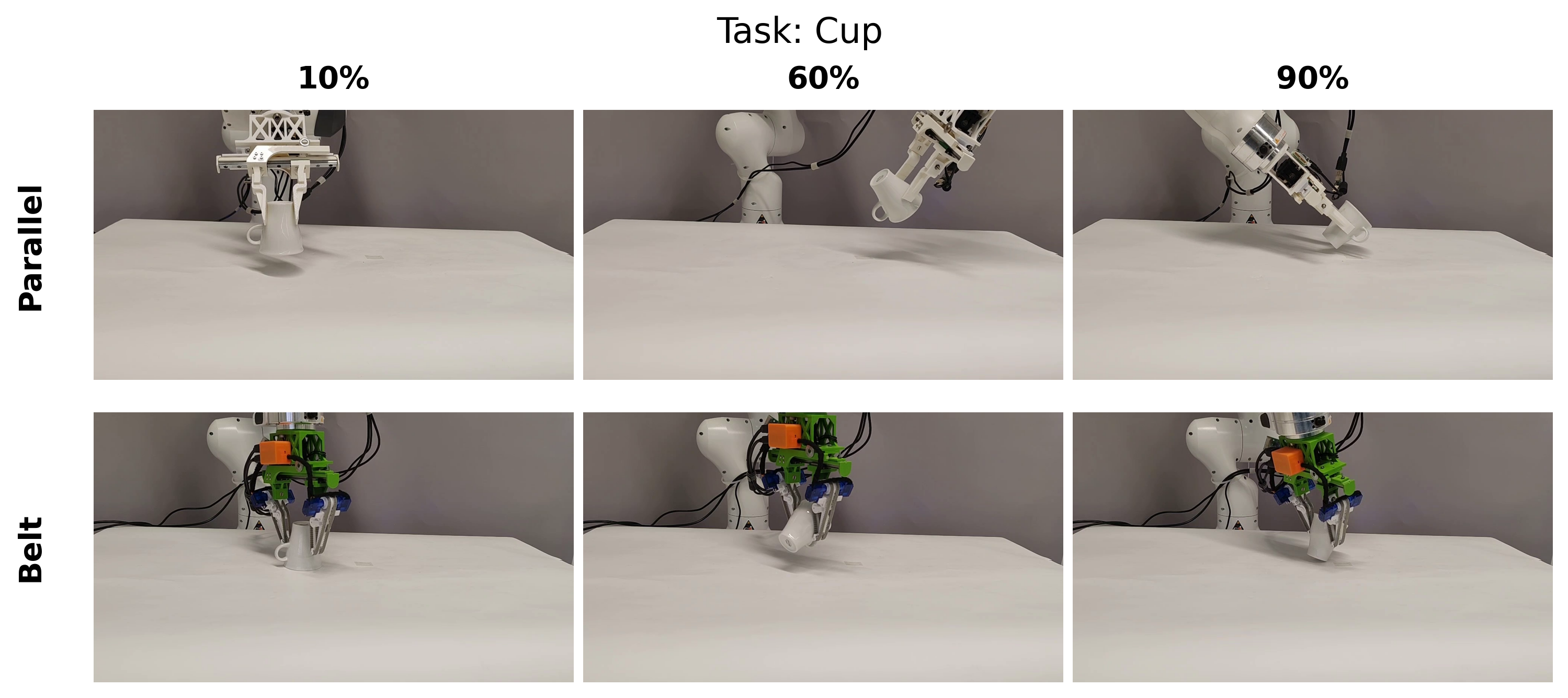}
\end{subfigure}
\begin{subfigure}{.49\linewidth}
  \includegraphics[width=\linewidth]{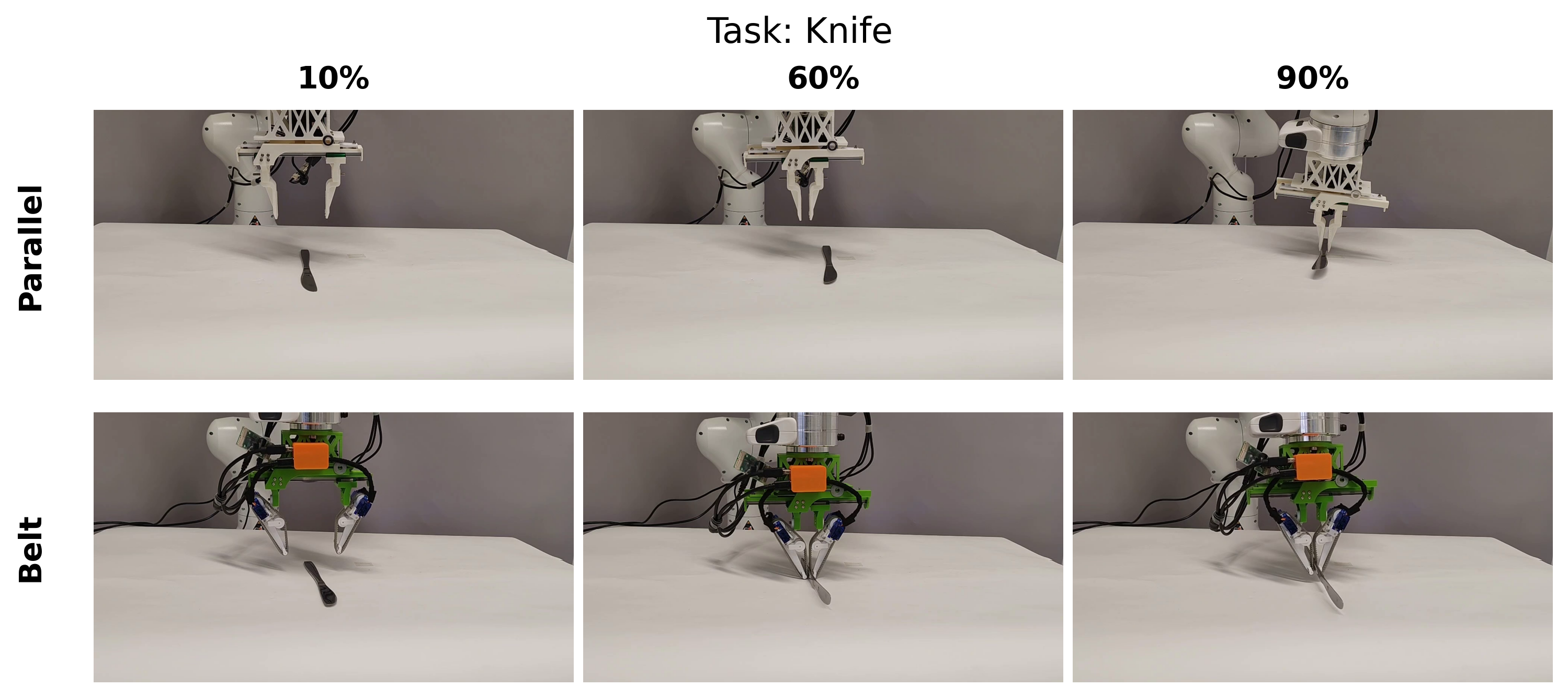}
\end{subfigure}
\begin{subfigure}{.49\linewidth}
  \includegraphics[width=\linewidth]{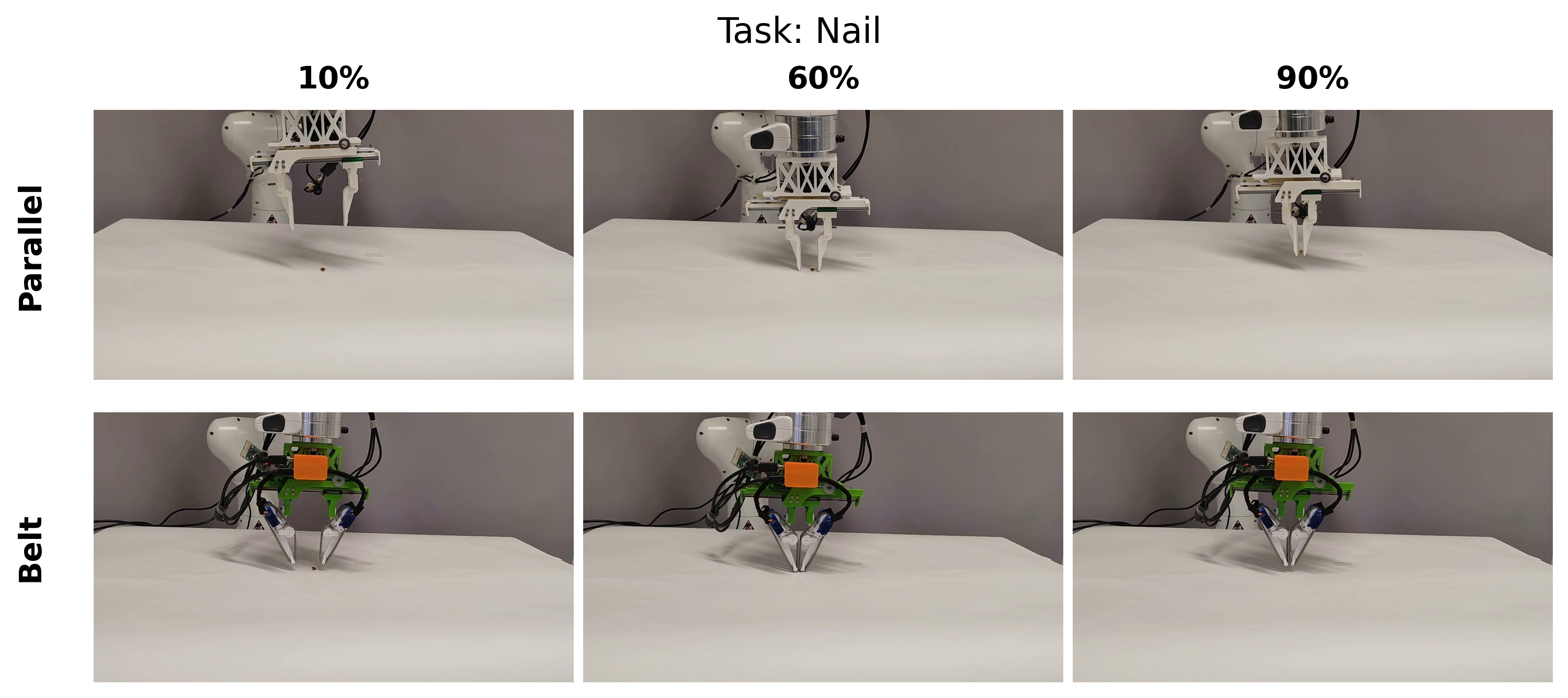}
\end{subfigure}
\begin{subfigure}{.49\linewidth}
  \includegraphics[width=\linewidth]{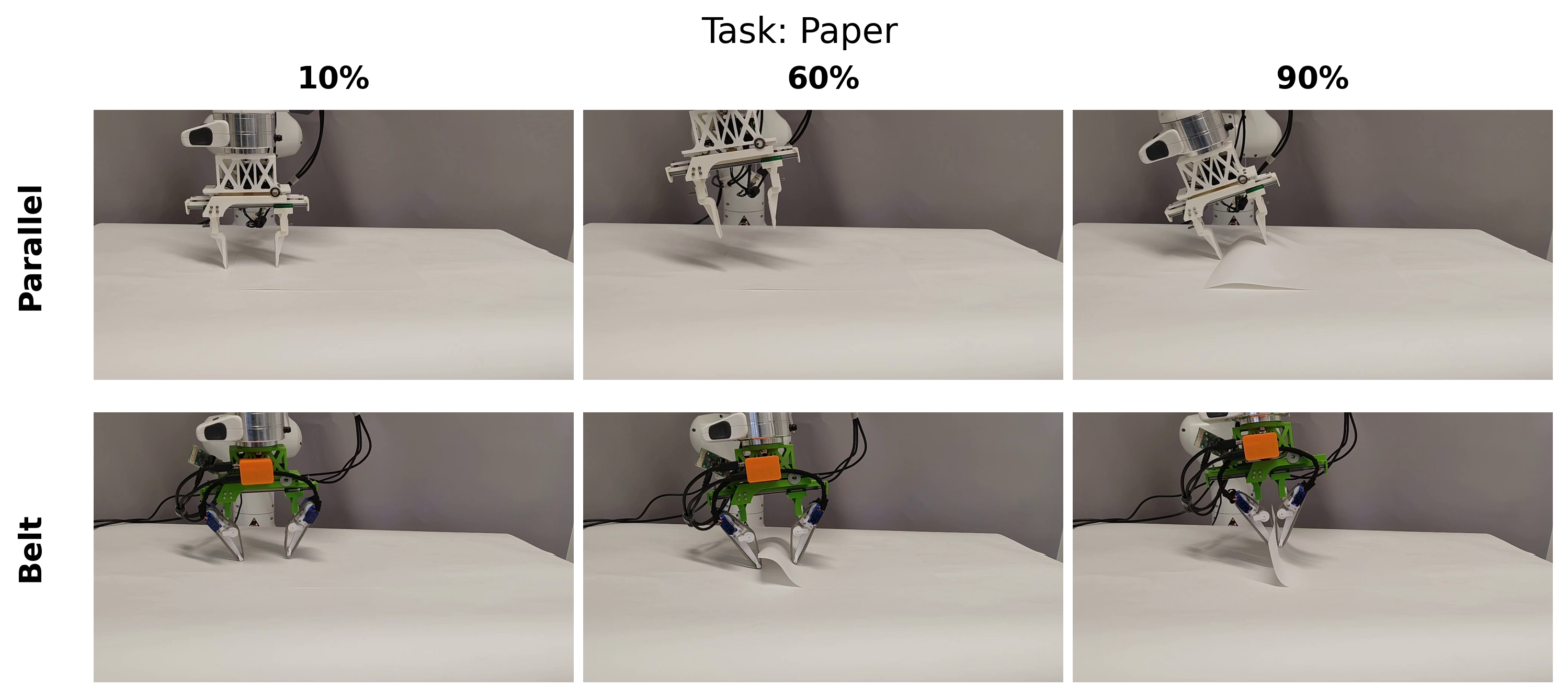}
\end{subfigure}
\begin{subfigure}{.49\linewidth}
  \includegraphics[width=\linewidth]{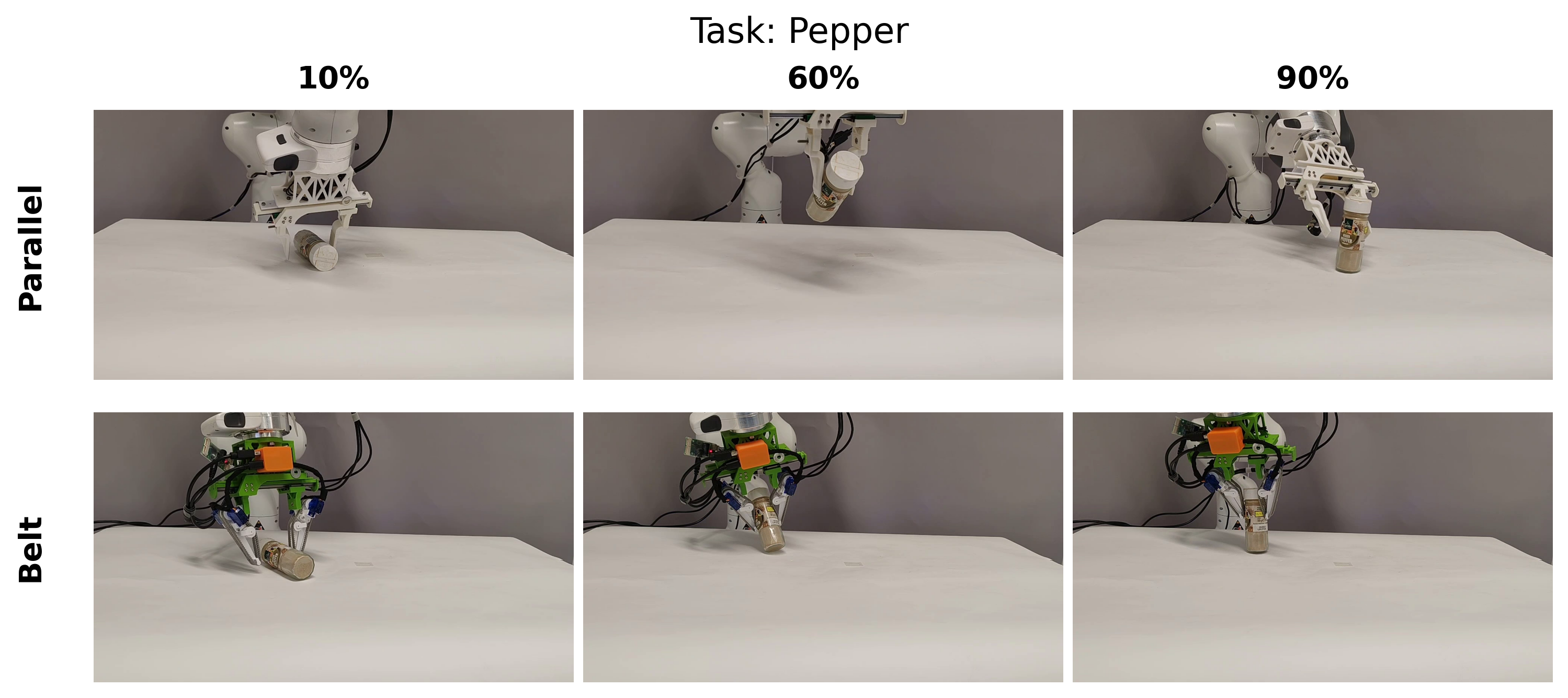}
\end{subfigure}
\begin{subfigure}{.49\linewidth}
  \includegraphics[width=\linewidth]{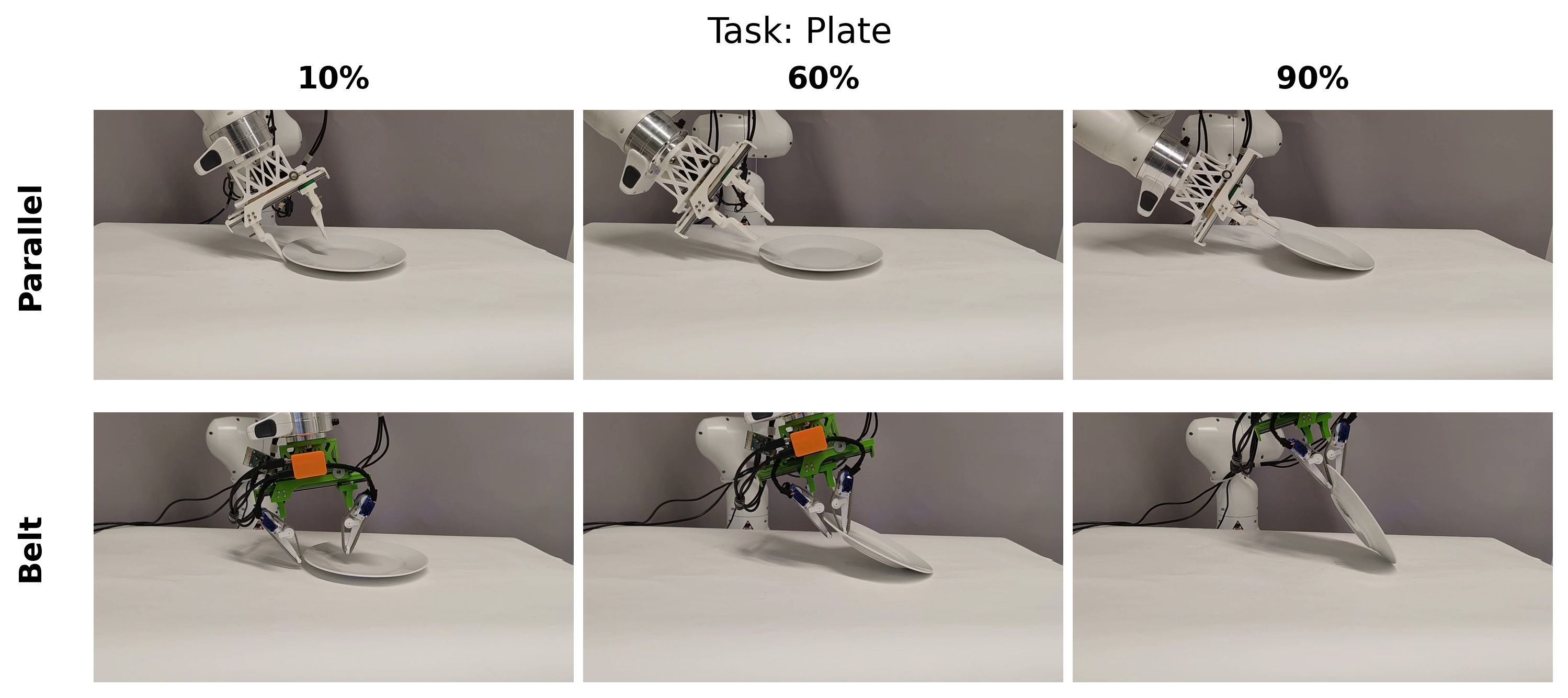}
\end{subfigure}
\begin{subfigure}{.49\linewidth}
  \includegraphics[width=\linewidth]{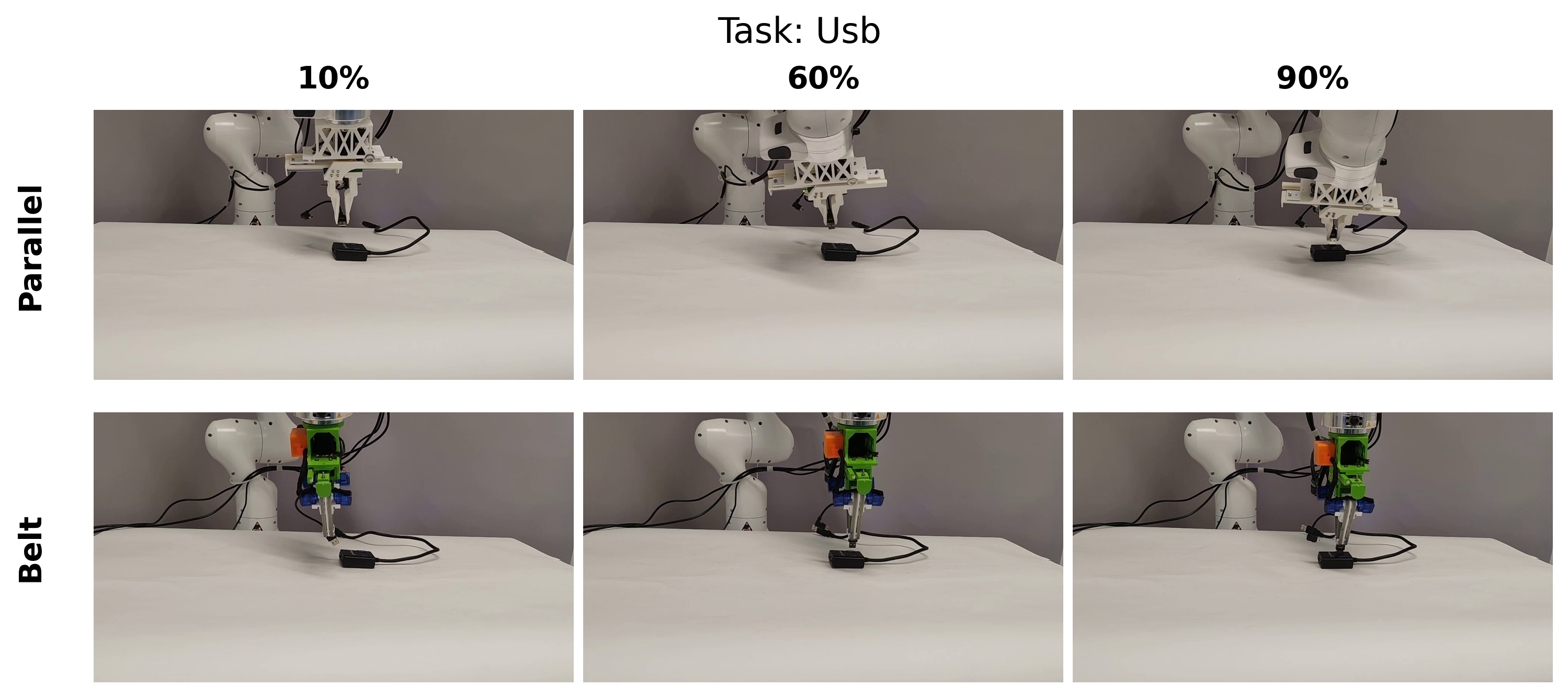}
\end{subfigure}
\hspace{.2\linewidth}

  \caption{Comparison between the conventional parallel finger with the \method{} on ten daily tasks. Task descriptions are listed in \cref{tab:task-description}}
  \label{fig:daily-task}
\end{figure*}

As shown in \cref{fig:daily-task}, the \method{} is capable of solving all ten tasks, while the parallel gripper failed at five tasks. Especially for tasks like Paper, Knife, and Allen, which require the in-hand pitch and roll motion, it is almost impossible to solve using a parallel gripper.
For tasks Bottle, Cloth, and USB, the \method{} has a much higher grasping stability. With in-hand reorientation and translation, tasks are solved much more efficiently.
For instance, the flipping motion usually requires more than two steps for a conventional parallel gripper, which instead can be done with a pitch motion for the \method{}.
By leveraging in-hand translation, \method{} addresses 'insufficient grasping'—typically caused by pinching the edges of small objects or fabric—by shifting the contact point further toward the center of the gripper.
Meanwhile, the complex screw operation can be performed by the roll motion.

\section{Gripper Durability and Long-Term Friction Changes}\label{app:friction}
Although comprehensive wear endurance testing falls outside the scope of our current study, we are pleased to present the second payload test results in \cref{tab:payload-result-rebuttal}.
After moderate usage, the belt shows a slight payload reduction with a mean decay of 7.5\%.
We designed the finger frame to carry the load during large deformations, preventing the Mullins effect in the belt.

\begin{table}
\centering
\caption{\textsc{Measured payloads of test objects}} \label{tab:payload-result-rebuttal}\vspace{.2em}
\vspace{-.2em}
The test was conducted after roughly 40 hours of moderate use (including data collection and demonstrations) following the original tests in the paper.
\vspace{.5em}
\adjustbox{max width=\linewidth}{
\begin{tabular}{@{}c|c|c|c}
\toprule
Object & Translation ($\mathrm{N}$) & Roll ($\mathrm{N\cdot mm}$)         & Pitch ($\mathrm{N\cdot mm}$) \\

\midrule
\multirow{2}{*}{cuboid} & $\!\!-10.66 \pm 1.95$ & $-114.6 \pm 13.87$ & $-55.54 \pm 6.04$ \\
                        & $+4.17 \pm 0.65$ & $+64.59 \pm 11.93$ & $+38.43 \pm 7.12$  \\\cline{2-4}
\multirow{2}{*}{cylinder} & $-5.87 \pm 0.81$ & $-54.96 \pm 4.19$ & $-29.42 \pm 2.97$ \\
                          & $+2.09 \pm 0.47$ & $+30.46 \pm 4.52$ & $+20.30 \pm 2.18$  \\ \cline{2-4}
\multirow{2}{*}{cone} & $\!\!-6.39 \pm 0.8$ & $-69.02 \pm 8.96$ & $-34.68 \pm 5.28$\\
                      & $+2.35 \pm 0.58$ & $+38.58 \pm 5.71$ & $+24.96 \pm 3.46$\\ \cline{2-4}
\multirow{2}{*}{sphere} & $-7.51 \pm 1.24$ & $-156.63 \pm 20.77$ & $-49.21 \pm 8.48$\\
                        & $+3.92 \pm 1.11$ & $+103.53 \pm 22.21$ & $+32.12 \pm 8.76$\\
\bottomrule
\end{tabular}
}
\vspace{-1.2em}
\end{table}

\section{MPC Control with Complex Object}\label{app:complex-object-manipulation}
\begin{figure}
\centering
  \includegraphics[width=.95\linewidth]{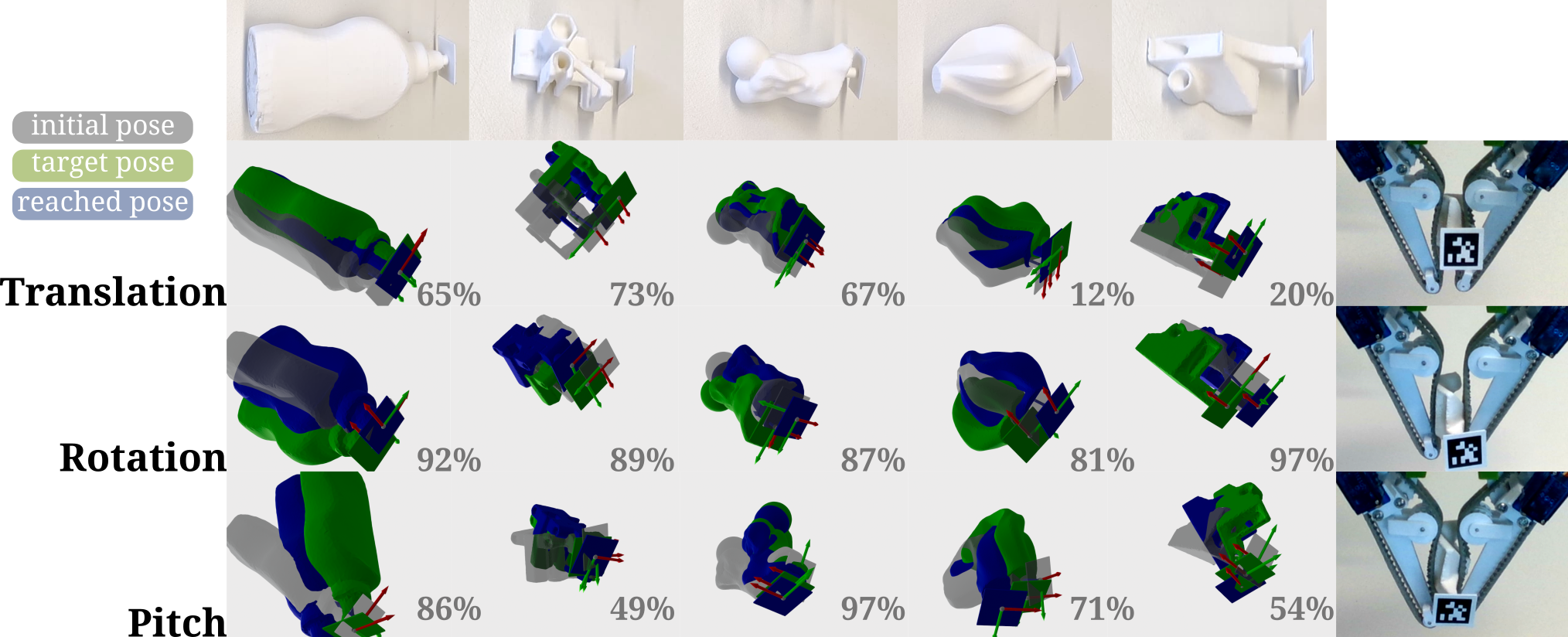}
  \caption{\textbf{In-hand motion for complex objects with MPC.} The percentage shows the achieved progress (For Translation: relative Cartesian distance. For Rotation and Pitch: relative angle)}
  \label{fig:mpc-object}
\end{figure}

To further demonstrate the capabilities of our soft belt and the proposed MPC controller, we conducted in-hand manipulation experiments using five complex objects 
as shown in \cref{fig:mpc-object}. 
While the complex surfaces of these objects sometimes prevent the system from reaching the exact final target pose, the MPC successfully manipulates most objects with high alignment. 
We evaluated the averaged pose errors using the translation and rotation error separately, achieving: $9.4 \pm 5 \text{mm}$,$14 \pm 16^\circ$\textbf{(translation)}, $14 \pm 4 \text{mm}$, $16 \pm 23 ^\circ$ \textbf{(roll)}, $13 \pm 6 \text{mm}$,$13 \pm 11 ^\circ$ \textbf{(pitch)}, $13 \pm 8 \text{mm}$, $33 \pm 25 ^\circ$ \textbf{(combi)}.
(Note that the magnitude of the translation error is largely an artifact of the distance metric relative to the radian-based rotation). 
Crucially, these complex in-hand rotations are entirely dependent on the wrapping effect of the soft belt; without it, such manipulation would be impossible.

\section{VLA Evaluation Details}\label{supp:vla}
We use the official implementation of GR00T N1.7, and the Lerobot implementation for $\pi{0.5}$ and ACT.
For ACT, we train one model for each task from scratch for evaluation.

All methods use the same observations: external \& wrist camera ($224 \times 224$), joint positions (7 dim), gripper opening (1 dim), end-effectors translation (3 dim), and rotation (6 dim). 
The image is un-normalized for $\pi{0.5}$. 
The rest of the observation is normalized using the quartile.
The action is increased to 10 dim: relative end-effector pose (6 dim), gripper opening (1 dim), gripper translation (1 dim), gripper roll (1 dim), gripper pitch (1 dim).
For each model, we fine-tune for a maximum of 100k steps.

During inference, we use a action chunk size of 20.
For ACT, we enable Temporal Ensembling.
The policies are evaluated using the same scripts with a different inference backend setup in a Singularity container running in an RTX 5090 GPU.
We use 20 Hz as the inference frequency.
A timeout of 30 sec is set as the truncation.

To evaluate the closed-loop control of the policy during in-hand reorientation and translation, we intentionally perturb the motor's command to the velocity ratio.
This effectively prevents the model from overfitting to the control command without aligning with the wrist camera visual information.

Evaluation clips can be found in the appended video.

\section{Model Predictive Control details}\label{app:MPC}

\subsection{Point contact approximation in MPC}

\textbf{The reason why having MPC:}
In-hand manipulation is essentially a path planning task for a given object.
The order of translations and rotations needs to be executed in the correct order to enable the motion, which can be pre-planned based on the gripper contact model.
For contact-rich tasks, the hybrid model predictive control is usually considered with mixed-integer programs \cite{hogan2020reactive}.
However, for the proposed belt finger, the contact points strongly depend on both the shape of the object and its friction coefficient distribution.
Meanwhile, the belt's elasticity and the dynamics of the gripper's open/close motion also affect the contacts in the real world.
Thus, long-horizon planning usually diverges rapidly from the real state with an inaccurate model.

To tackle this issue, we separately consider the belt motion and the open/close motion.
In the proposed MPC algorithm, the short-horizon planning uses iCEM only for belt action.
On the other hand, the gripper motion is actively adapted to the belt motion to achieve the desired object movement.
Thus, an extra re-estimation of contact points is used to determine the real contact states to calculate the gripper's next open/close motion.
This method is simple but also effective with an inaccurate model.

\subsection{The general idea of re-evaluation of contact points (RCP)}\label{app:mpc:pressure}

In hybrid MPC, to determine the current mode of the MPC, we need to know the number and location of the contact points (approximated contact points).
Since no tactile feedback is directly given from the observation, the theoretical contact point is estimated based on the assumption in \cref{sec:model} and the object pose for each MPC planning.

To maintain the contact mode estimated by the MPC, the open/close motion of the gripper needs to adequately coordinate with the planned belt motion.

Although it is possible to optimize the gripper movement together with the belt motion using a zero-order planner, due to the curse of dimensionality during the sampling process, it is reasonable to estimate the open/close motion separately.

As shown in \cref{alg:mpc}, the contacts $\hat{C}^{t}$ are estimated based on the current object pose $o^{t}$ and the gripper opening size $a^t_{opening}$ through their collision model.
For MPC planning, the estimated contacts $\hat{C}^{t}$ are used for the model rollout.
After the execution of the proposed action $a^t$, the re-evaluation of the previous contacts $\hat{C}^{t}$ is performed to determine whether the estimated contact points produce valid contacts or fake contacts, e.g., slip.
To do so, we compare the distances between the real reached object pose $o^{t+1}$ and theoretical target object poses $\hat{o}^{t+1}$  with all possible slip combinations ($P(\hat{C}^{t})$).

The contact with the closest distance is the hindsight contact $C^t_h$. The corresponding target pose is the hindsight pose $\hat{o}^{t+1}_{C_h}$.

To determine if the pressure between the fingers matches the contact mode, two criteria are used:
1. It is over-pressure if the object hasn't moved.
2. It is under-pressure if the number of contacts after the re-evaluation ($C^t_h$) is smaller than the previous estimated one ($\hat{C}^t$).
The next gripper opening size $l^{t+1}_{\delta}$ will be adapted by  $\Delta l$ to increase, maintain, or decrease the pressure accordingly for the next motion.

\subsection{Details on the \method{} model}\label{app:mpc:model}
Here, we detail how the object pose and velocity are updated in our simplified model.
We start from the contact points  $c^t_i$ estimated using ray tracing\cite{trimesh} at time step $t$. We then compute relative positions given by $r^t_{i,j}=c^t_j-c^t_i$ and relative velocities given by $v^t_{i,j}=v^t_{c_j}-v^t_{c_i}$, $[\cdot]_{\times}$ represents the twist matrix, and $[\cdot]^{\dagger}$ denotes its pseudo-inverse. All calculations are in the gripper frame.
\begin{equation}
\begin{split}\label{eq:mpc-vel}
\omega^t_o
&=
-1
\cdot
\begin{bmatrix}[r^t_{1,3}]_{\times}\\ [r^t_{1,4}]_{\times}\end{bmatrix}^{\dagger}
\cdot
\begin{bmatrix}v^t_{1,3}\\ v^t_{1,4}\end{bmatrix} \\
v^t_o
&=
\frac{1}{3}\sum_{i\in\{1,3,4\}}
v^t_{i} - [\omega^t_o]_{\times}
\cdot
(c^t_i-o^t)
\end{split}
\end{equation}

When the object has only two active contact points, we remove the non-contact term from the \cref{eq:mpc-vel}.
For the single active contact scenario, we use a translation-only assumption.
This means that the object only conducts translation motion due to the stickiness of the contact point.
If no active contacts are presented, only gravity is driving the object's motion.

The estimated pose $\hat{o}^{t+1} = \begin{bmatrix}\hat{R}^{t+1} & \hat{T}^{t+1}\\ \mathbf{0} & 1 \end{bmatrix}$ can be calculated by one step integration for the full contacts scenario in \cref{eq:mpc-vel-integration}.

\begin{equation}
\begin{split}\label{eq:mpc-vel-integration}
\hat{R}^{t+1} &= \exp([\omega_o^t dt]_{\times})\cdot R^{t} \\
\hat{T}^{t+1} &= v^t_o dt+ T^{t}
\end{split}
\end{equation}


\section{Discussion of Experiment \cref{sec:ex2}}\label{app:ex5}

For in-hand manipulation with limited DoFs, not all target poses are reachable.
Target poses with four different difficulties are proposed for the MPC controller.
In the Translation case, only one translation is necessary to reach the target pose.
In the Easy Combi case, both rolling and pitching motions are necessary to reach the target pose.
In the Medium Combi case, all three motions are necessary to reach the target pose.
In the Impossible Combi case, the target pose is not reachable due to the parallel gripper. The major reason to include this is to investigate to what extent does the proposed MPC algorithm performs in the worst case.

The pose error of each target difficulty is listed in \cref{tab:pose-error-mpc} using the canonical left metric in $SE(3)$ \cite{JMLR:v21:19-027} as the distance metric.


It is also worth pointing out that although each target pose can be decomposed into translation and rotation in the gripper frame, this doesn't guarantee a valid manipulation trajectory. Especially when the belt actions are not strictly mapping to the rotation or translation of the object due to the friction and the flexibility of the belt.

\section{Design Choices} \label{app:design-choices}
\begin{figure}
\centering
\begin{minipage}[t]{.4\linewidth}
  \includegraphics[width=\linewidth]{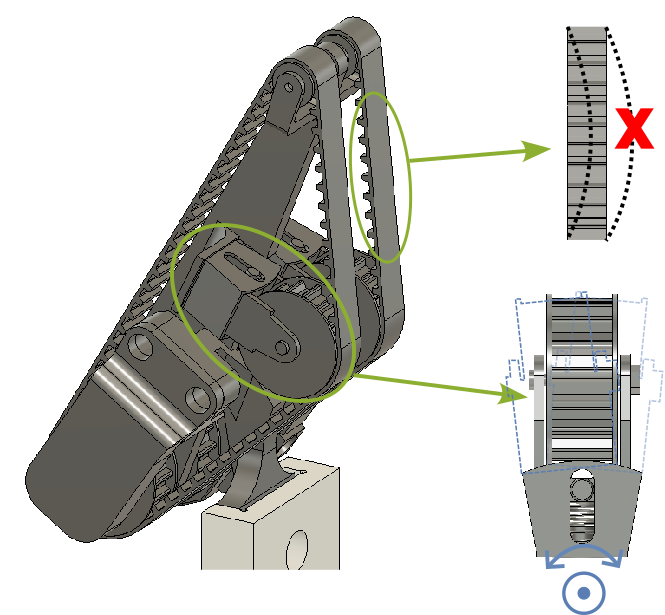}
  \caption{Rotatable transmission structure. This design released the rotation DoF around the z-axis. However, the extra motion freedom reduces the belt stability compared to the simpler design.}
  \label{fig:rotable-structure}
\end{minipage}
\end{figure}

To achieve a better wrapping around the grasped object, the belt should be flexible transversely and longitudinally.
However, the belt per se is an approximated ruled surface mainly due to the existence of the tooth, which works like stiffeners, preventing it from bending transversely, as shown in \cref{fig:rotable-structure}.

\begin{figure}[ht]
  \centering
  \begin{minipage}[t]{0.48\linewidth}
    \centering
    \vspace{0pt} 
    \includegraphics[width=\linewidth]{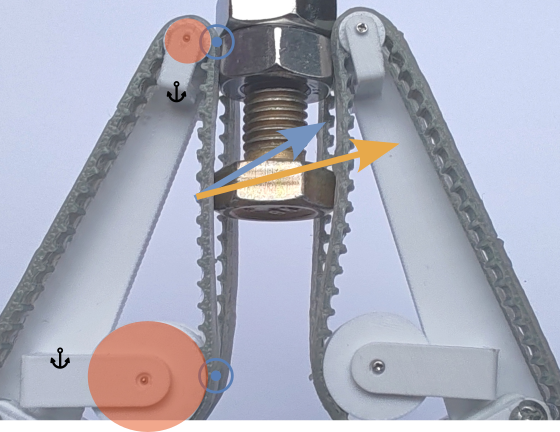}
    \caption{Fixed wheel in the simpler design results in a less complete wrapping when the non-convex object has multiple contact areas along the belt.}
    \label{fig:non-parallel-wrap}
  \end{minipage}
\begin{minipage}[t]{0.35\linewidth}
\centering
\vspace{0pt} 
\includegraphics[width=\linewidth]{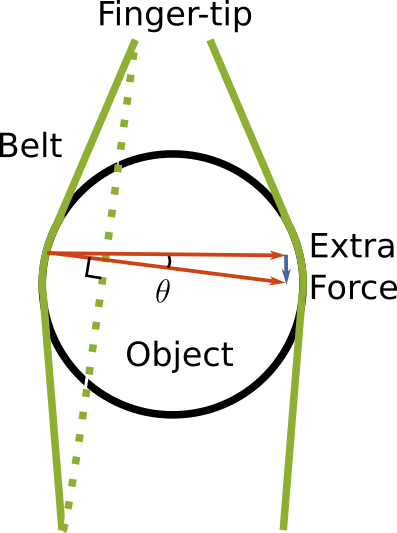}
\caption{Non-parallel belt design to reduce motor load during translation lifting.}
\label{fig:belt-gripper-z-force-issue}
\end{minipage}
\end{figure}

When a pair of fixed wheels drives the belt, all twists around the longitudinal axis need to be re-twisted between the fixed ends.
This reduces the wrap-around capability of a belt when the contact surface of the grasped object is not perfectly parallel to the neutral belt position, as shown in \cref{fig:non-parallel-wrap}.


An attempted solution to this problem, which proved unsuccessful, is to release one DoF of the fixed wheel, as shown in \cref{fig:rotable-structure}. The middle wheel can rotate transversely along the belt, which allows for a better fit of the object.

However, the extra shear force in the contact surface pushes the belt further sideways, which is less of an issue when the arc length is smaller with fixed wheels. Thus, the design with fixed wheels is adopted.

As shown in \cref{fig:belt-gripper-z-force-issue}, the belts of the antipodal fingers are non-parallel, with an angle $2\theta$. This induces an extra lifting force from the gripper closing force $F_{close}\sin{\theta}$.
Firstly, this special design produced pointed finger-tips that reduce the contact area of the belts when picking small objects.
Secondly, the extra lifting force alleviates motor strain during the translational movement of heavy loads.
As described in \cref{fig:payload-ex-setup}, the gripper provides up to 9N translation force for downward motion, and 4 N forces upwards.
This non-symmetric force measure is directly caused by this special design.

\clearpage
\acknowledgments{If a paper is accepted, the final camera-ready version will (and probably should) include acknowledgments. All acknowledgments go at the end of the paper, including thanks to reviewers who gave useful comments, to colleagues who contributed to the ideas, and to funding agencies and corporate sponsors that provided financial support.}



\end{document}